\PassOptionsToPackage{table}{xcolor}

\documentclass[acmtog]{acmart} 

\acmSubmissionID{1525}

\AtBeginDocument{%
  }

\copyrightyear{2025}
\acmYear{2025}
\setcopyright{acmlicensed}\acmConference[SA Conference Papers '25]{SIGGRAPH Asia 2025 Conference Papers}{December 15--18, 2025}{Hong Kong, Hong Kong}
\acmBooktitle{SIGGRAPH Asia 2025 Conference Papers (SA Conference Papers '25), December 15--18, 2025, Hong Kong, Hong Kong}
\acmDOI{10.1145/3757377.3763888}
\acmISBN{979-8-4007-2137-3/2025/12}

\usepackage{cleveref}

\usepackage{soul} 
\sethlcolor{orange} 

\usepackage[table]{xcolor}
\usepackage{subcaption}
\usepackage{makecell}  
\usepackage{colortbl}

\usepackage{enumitem}

\begin{document}

\title{Virtually Being: Customizing Camera-Controllable Video Diffusion Models with Multi-View Performance Captures}

\author{Yuancheng Xu}
\orcid{0000-0002-2254-5752}
\affiliation{\institution{Eyeline Labs}
\country{United States of America}}
\email{xuyuancheng0@gmail.com}

\author{Wenqi Xian}
\orcid{0009-0007-4678-6458}
\affiliation{\institution{Eyeline Labs}
\country{United States of America}}
\email{wenqixian3@gmail.com}

\author{Li Ma}
\orcid{0000-0002-6992-0089}
\affiliation{\institution{Eyeline Labs}
\country{United States of America}}
\email{lmaag@connect.ust.hk}

\author{Julien Philip}
\orcid{0000-0003-3125-1614}
\affiliation{\institution{Eyeline Labs}
\country{United Kingdom}}
\email{julien.philip@scanlinevfx.com}

\author{Ahmet Levent Taşel}
\orcid{0009-0002-7150-0160}
\affiliation{\institution{Eyeline Labs}
\country{Canada}}
\email{leventtasel@gmail.com}

\author{Yiwei Zhao}
\orcid{0000-0001-6276-8021}
\affiliation{\institution{Netflix}
\country{United States of America}}
\email{yiweiz@netflix.com}

\author{Ryan Burgert}
\orcid{0009-0008-5947-2076}
\affiliation{\institution{Eyeline Labs}
\country{United States of America}}
\email{ryancentralorg@gmail.com}

\author{Mingming He}
\orcid{0000-0002-9982-7934}
\affiliation{\institution{Eyeline Labs}
\country{United States of America}}
\email{hmm.lillian@gmail.com}

\author{Oliver Hermann}
\orcid{0009-0000-6852-2495}
\affiliation{\institution{Eyeline Labs}
\country{Germany}}
\email{oliver.hermann@scanlinevfx.com}

\author{Oliver Pilarski}
\orcid{0009-0005-7803-2314}
\affiliation{\institution{Eyeline Labs}
\country{Germany}}
\email{oliver.pilarski@scanlinevfx.com}

\author{Rahul Garg}
\orcid{0009-0003-6435-9346}
\affiliation{\institution{Netflix}
\country{United States of America}}
\email{rahulgarg@netflix.com}

\author{Paul Debevec}
\orcid{0000-0001-7381-2323}
\affiliation{\institution{Eyeline Labs}
\country{United States of America}}
\email{debevec@gmail.com}

\author{Ning Yu}
\orcid{0009-0004-6865-1325}
\affiliation{\institution{Eyeline Labs}
\country{United States of America}}
\email{ningyu.hust@gmail.com}

\renewcommand{\shortauthors}{Xu et al.}

\begin{abstract}
We introduce a framework that enables both multi-view character consistency and 3D camera control in video diffusion models through a novel customization data pipeline. 
We train the character consistency component with recorded volumetric capture performances re-rendered with diverse camera trajectories via 4D Gaussian Splatting (4DGS), lighting variability obtained with a video relighting model. 
We fine-tune state-of-the-art open-source video diffusion models on this data to provide strong multi-view identity preservation, precise camera control, and lighting adaptability. 
Our framework also supports core capabilities for virtual production, including multi-subject generation using two approaches: joint training and noise blending, the latter enabling efficient composition of independently customized models at inference time; 
it also achieves scene and real-life video customization as well as control over motion and spatial layout during customization.
Extensive experiments show improved video quality, higher personalization accuracy, and enhanced camera control and lighting adaptability, advancing the integration of video generation into virtual production.
Our project page is available at: \url{https://eyeline-labs.github.io/Virtually-Being/}.

\end{abstract}


\begin{teaserfigure}
  \centering
  \includegraphics[width=\textwidth]{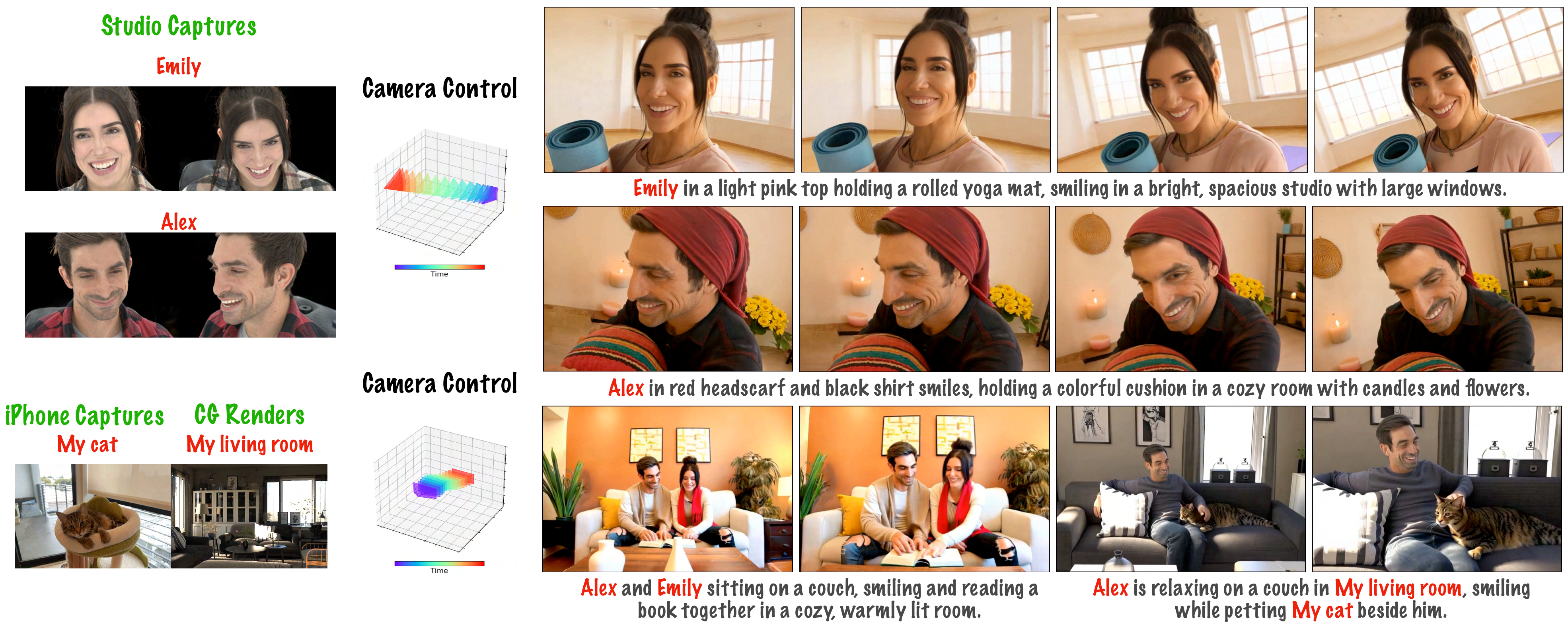}
  \caption{Leveraging multi-view data from 4D reconstructions from studio captures, CG renders, and real-life videos, our method enables video generation with strong multi-view identity preservation and camera control, while also supporting multi-subject composition and subject–scene interactions. }
  \label{fig:teaser}
\end{teaserfigure}
\maketitle

\section{Introduction}
\label{sec:introduction}

Virtual production is reshaping filmmaking by combining live-action footage with computer-generated elements through technologies such as LED volumes, camera tracking, and performance capture. This integration allows filmmakers to interact with virtual environments during production, offering direct control over scene layout, lighting, character performance, and camera motion. Increasingly, video generation models~\cite{blattmann2023align,blattmann2023stable,guo2024animatediff,yang2024cogvideox,kong2024hunyuanvideo,wan2.1} are being incorporated into this pipeline to further expand creative possibilities and streamline content creation. 
A key application is subject-specific customization, enabling the generation of novel scenes featuring particular characters.
Additionally, in cinematic storytelling, camera motion is essential for conveying perspective and emotion, but it also exposes the subject from multiple angles, posing a significant challenge for preserving identity. As a result, multi-view identity preservation becomes crucial for generating coherent and realistic customized videos with camera control.

However, despite growing interest in customized video generation, explicit multi-view identity preservation under camera motion remains largely unexplored. Recent methods often customize subjects using a single reference image~\citep{yuan2024identity, jiang2024videobooth, chen2025multi}, which lacks view diversity and leads to identity inconsistencies when the subject is observed from novel angles. Moreover, existing approaches offer limited camera control during customization. For example, MotionBooth~\citep{wu2024motionbooth} supports only 2D camera translations, falling short for 3D camera movements common in filmmaking.

In this work, we introduce an approach that enables both multi-view subject customization and 3D camera control by repurposing 4D Gaussian Splatting (4DGS) from volumetric captures~\citep{yang2023gs4d} — originally designed for 4D reconstruction — as a data generator for video diffusion models. This data pipeline bridges 4DGS’s accurate multi-view rendering with the generative flexibility of video generation models. 
We begin by capturing dynamic human performances using two professional volumetric capture rigs: a 75-camera facial setup and a 160-camera full-body system. From these captures, we apply 4DGS to reconstruct dynamic human motion and render videos with diverse and precisely annotated camera trajectories. To reflect the central role of lighting in cinematography, we further augment the data using a generalizable relighting model~\cite{mei2025lux}, producing HDR-based lighting variations. This pipeline provides rich multi-view character supervision, accurate camera conditioning, and lighting diversity essential for high-fidelity generation. Building on this customization dataset, we adopt a two-stage training strategy: first pretraining on general camera-annotated datasets for camera-conditioned synthesis, then fine-tuning on our customized data to achieve multi-view identity preservation under 3D camera motion.

Our framework supports a range of features essential for virtual production, enabling flexible control over subjects, scenes, and actions in generated videos. (1) Multi-subject Generation: We enable multi-subject video generation by jointly customizing the model on subject-specific datasets. In addition to joint training, we incorporate a noise blending approach~\cite{kong2024omg} that combines independently customized models at inference time by blending LoRA features using segmentation masks from GroundingDINO~\cite{liu2024grounding} and SAMv2~\cite{ravi2024sam}, enabling efficient subject composition. (2) Scene Customization: Our method supports scene-specific generation using high-quality CG videos curated by professional artists, allowing novel subject–scene interactions under diverse camera motions. (3) Real-life Customization Data: Beyond 4DGS data, we validate our pipeline on real-life videos with estimated camera parameters~\cite{wang2025continuous}, demonstrating its effectiveness in real-world settings. (4) Action and Spatial Layout Control:  By fine-tuning the Go-with-the-Flow model~\citep{burgert2025go} on our customized dataset, we enable control over subject motion and layout, preserving spatial arrangement and movement patterns from source videos.

In summary, our key contributions are as follows:
\begin{enumerate}[topsep=0pt]
    \item We present the first framework to explicitly preserve multi-view identity under precise 3D camera control, enabled by a novel customization data pipeline that integrates professional volumetric captures, 4DGS reconstruction, and relightable rendering — combined with a two-stage training strategy that first learns general camera-conditioned video generation and then customizes to specific subjects.
    \item Our method supports a broad range of generative capabilities for filmmaking, including multi-subject generation via both joint training and a noise blending scheme for combining independently customized models, scene customization, real-life video-based customization, and control over subject motion and spatial layout.
    \item We conduct extensive benchmarking, ablations, and user studies, demonstrating clear improvements in multi-view identity preservation, camera control accuracy and lighting control—highlighting the method’s utility for virtual production applications.
\end{enumerate}

\section{Related work}
\label{sec:related_work}

\subsection{Scene and character customization}

Early video diffusion models, while capable of generating content from text prompts, often struggled with temporal consistency and multi-view coherence.  More recent approaches address these issues by incorporating structured scene representations.  Methods like ReconX \cite{liu2024reconx}, ViewCrafter \cite{yu2024viewcrafter}, and Wonderland \cite{liang2024wonderland} combine diffusion models with 3D representations like point clouds or 3D Gaussian Splatting (3DGS) \cite{kerbl3Dgaussians}. However, these approaches often struggle with fast scene motion , and typically focus only on scene generation with less explicit control over characters in the scene.

Character customization in video diffusion models can be approached through finetuning-free or finetuning-based methods. Fine-tuning-free methods, such as Videobooth~\cite{jiang2024videobooth} and ConsisID~\cite{huang2024consistentid}, generate subject-consistent videos from a single reference image, while ConceptMaster~\cite{huang2025conceptmaster}, VideoAlchemist~\cite{chenvideoalchemy}, and Phantom~\cite{liu2025phantom} extend these ideas to multi-subject video generation. However, in scenarios involving subject motion and camera motion, where viewpoints can vary significantly, a single reference image often fails to capture the full multi-view characteristics of the subject. Fine-tuning-based methods, including DreamVideo~\cite{wei2024dreamvideo}, VideoStudio~\cite{long2024videostudio}, Magic-Me~\cite{ma2024magic}, and \citet{yuan2024identity}, build on DreamBooth~\cite{ruiz2023dreambooth} to adapt pre-trained models to specific characters using small sets of images. \citet{chen2025multi} further extends fine-tuning approaches to multiple identities. 
These methods are often constrained by limited input data, reducing viewpoint diversity, and struggle to integrate customized characters into dynamic, consistently lit scenes with controlled camera motion. We address this with a fine-tuning-based strategy for both character and scene customization with controllable camera motions.

\subsection{Camera and lighting control}

Controlling camera movement is crucial for cinematic storytelling.  Early approaches used fine-tuning techniques, like AnimateDiff's \cite{guo2024animatediff} use of LoRAs~\cite{hu2021loralowrankadaptationlarge}, to manage specific motion types.  However, this offers limited precision.  More recent methods, such as MotionCtrl~\cite{wang2024motionctrl} and its successors \cite{he2024cameractrl,kuang2024collaborative,xu2024camco,he2025cameractrl}, condition video generation directly on camera extrinsics, often using Plücker embeddings \cite{sitzmann2021light}.  Training-free strategies \cite{ling2024motionclone, xiao2024video, hou2024training} exist but can be difficult to integrate with other control aspects.  4D scene generation methods \cite{wu2024cat4d, watson2024controllingspacetimediffusion, sun2024dimensionx} offer inherent camera control, but their synthesis quality currently lags behind dedicated video generation models. Our approach builds upon AC3D~\cite{bahmani2025ac3d} and VD3D \cite{bahmani2025vd3d}, which effectively combine ControlNet \cite{zhang2023adding} and Plücker embeddings for high-quality, but extends this by adapting the camera conditioned model for other controls. 
Other works focus on redirecting camera trajectories in monocular videos~\citep{bai2025recammaster, yu2025trajectorycrafter}, while we aim to generate novel customized videos with controllable camera motion.
Most relevant to our work, MotionBooth~\cite{wu2024motionbooth} supports customized video generation with camera control but is limited to 2D translations, while our method handles full 3D camera motion.

Lighting control is another vital aspect of cinematography. While some methods allow for lighting adjustments via text prompts \cite{zhang2025scaling}, this lacks the precision required for professional cinematography.  DiffRelight \cite{he2024diffrelight} focuses on portrait relighting, which is limited in scope. \textit{Lux Post Facto} \cite{mei2025lux} provides state-of-the-art video-based relighting via HDR map encoding, which we use out of the box to augment our customization dataset with diverse lighting conditions.

\subsection{Controllable video generation with volumetric captures}

While the above methods control video generation by deploying the data prior in the video diffusion model, which is trained on millions of in-the-wild videos, another line of work focuses on acquiring high dimensional data with volumetric light stages and uses complex data processing pipeline to achieve fine-grained control for the final video. With synchronized multi camera settings, one can reconstruct 3D or 4D representations such as mesh \cite{DBLP:journals/tog/BeelerHBBBGSG11,DBLP:conf/eccv/CagniartBI10,DBLP:journals/cgf/FyffeHWMD11}, NeRF \cite{DBLP:journals/tog/LombardiSSSLS19,DBLP:journals/tog/IsikRGKSAN23} or Gaussian Splatting \cite{luiten2023dynamic,he2024diffrelight,Jiang_2024_CVPR}, which achieves camera controls. Further more, by designing 3D deformable model \cite{qian2024gaussianavatars,ma2024gaussianblendshapes} or skinned model \cite{peng2023implicit,peng2021neural}, one can achieve expression and pose control by providing detailed expression code and pose parameters. With data that captures the performance under different lighting conditions such as one-light-at-a-time (OLAT), photometric relighting can be achieved \cite{mei2025lux}. 
In this work, we leverage volumetric captures for customizing video generation models. 

\section{Method}
\label{sec:method}

\begin{figure}[htp!]
    \centering 
    \includegraphics[width=\columnwidth]{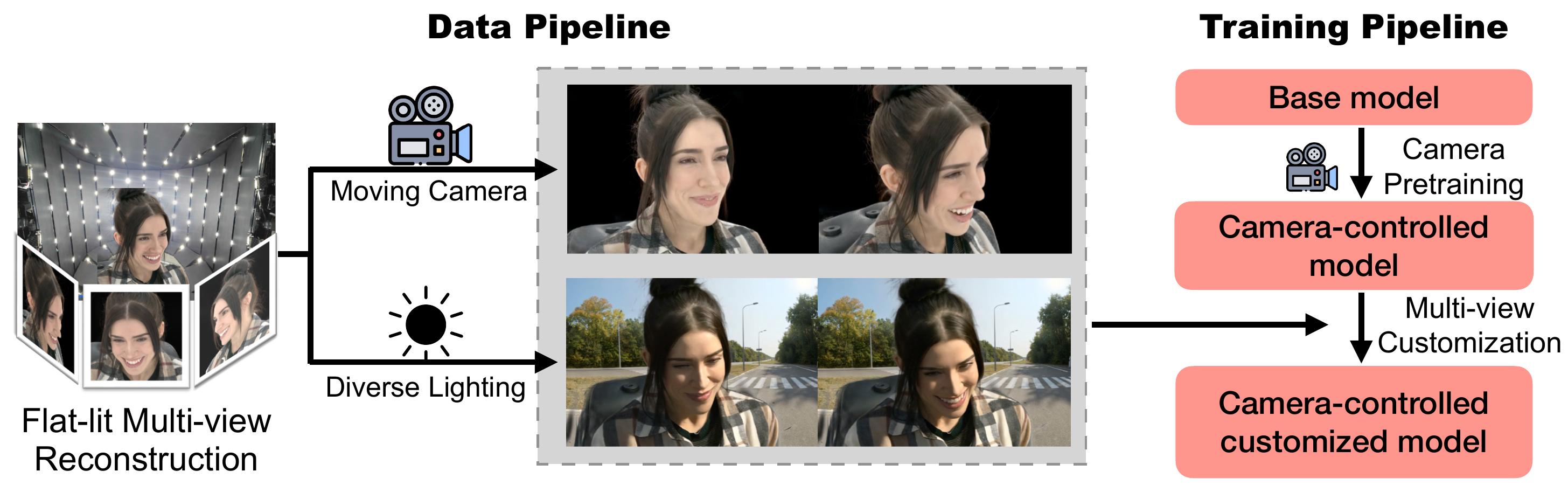}
    \caption{
    Overview of our training pipeline (right) for camera-controlled customized video generation, consisting of a camera pretraining stage and a customization stage. The data pipeline (left) generates customization data by capturing multi-view performances, applying 4D Gaussian Splatting, and rendering videos with diverse viewpoints, camera motions, and lighting.
    }
    \label{fig:workflow}
\end{figure}

\paragraph{Two-stage Training. } 
We aim to generate subject-specific videos that preserve multi-view identity across novel contexts with controllable camera motion. As viewpoint variation becomes more pronounced with subject and camera movement, maintaining multi-view consistency is critical. To address this, we adopt a two-stage training pipeline for both text-to-video and image-to-video models. In the pretraining stage, we train the model on diverse general datasets to learn camera-conditioned video generation across varying scenes and motions. In the customization stage, we fine-tune the model on subject-specific multi-view data to capture identity details and view-dependent variations. This design enables the model, at inference time, to generate identity-consistent, camera-controllable videos of customized subjects in diverse and unseen contexts.

\paragraph{Camera Pretraining Stage.}
To enable camera-controlled video generation of general content, we adopt a ControlNet architecture inspired by AC3D~\citep{bahmani2025ac3d}, where camera information is represented using Plücker coordinates~\citep{he2024cameractrl, kuang2024collaborative}. The camera representations are processed through a fully convolutional encoder, concatenated with video tokens from the main DiT, and passed through the ControlNet's DiT blocks. We integrate the camera information into the main DiT via token-wise summation before each block. Following AC3D, camera conditioning is applied only during the first 40\% of denoising timesteps and injected into the first 25\% of DiT blocks to improve controllability and visual quality. During training, we freeze the main DiT and train only the ControlNet on RealEstate10K~\citep{zhou2018stereo} for static scenes and HumanVid~\citep{wang2024humanvid} for dynamic, human-centric videos. The resulting model generates diverse general-content videos while accurately following specified camera trajectories.

\paragraph{Customization Stage.}
In this stage, we customize the camera-conditioned model to generate the target subject while preserving multi-view identity consistency, which is particularly crucial under camera motion. We adapt the DreamBooth framework~\citep{ruiz2023dreambooth}, fine-tuning the model on a subject-specific customization dataset while simultaneously using a regularization dataset—sampled from the pretraining data—to maintain general video generation and camera control capabilities. During customization, each subject is associated with a unique token embedded in the text prompts, allowing the model to learn to generate that specific subject when conditioned on the token. In the next section we detail how to construct effective customization datasets. 

\paragraph{Image-to-video Generation.}
In addition to customizing the video generation model, image-to-video (I2V) generation requires a subject-specific initial frame, which we generate using text-to-image (T2I) methods.
For single-subject generation, we fine-tune FLUX.1-dev using DreamBooth~\citep{ruiz2023dreambooth} on subject-specific images from our customization datasets. For multi-subject generation, we use MuDI~\citep{jang2024identity}, which extends FLUX.1-dev to support identity-consistent multi-subject synthesis. These customized images are then converted into videos using an camera-conditioned I2V model, which is also fine-tuned on the same customization dataset to preserve identity and support controllable motion. 

\subsection{Constructing customization datasets}

A key challenge in the customization stage is constructing an effective customization dataset: it must provide high-quality multi-view captures of the subject and precise camera annotations to support accurate camera control during generation. 
In the following, we discuss three sources of data: 4DGS from professional volumetric captures for human subject customization, CG scenes for scene customization and real-life videos. 

\paragraph{Professional Volumetric Captures.}
To obtain high-quality 4D reconstruction of a human subject, we first capture multi-view data in a controlled studio environment, using a volumetric face rig equipped with 75 synchronized cameras arranged on a cylindrical structure measuring 2.5 meters in height and 2.7 meters in diameter. Full-body performances are recorded using a larger body rig with 160 synchronized cameras mounted on a 4-meter-wide cylinder. Subjects are illuminated with multiple strobe lights to ensure flat and diffuse lighting. Each subject performs 3–6 multi-view sequences, each lasting approximately 50 to 180 frames at 24 frames per second. The captured subjects, referred to as “Alex” and “Emily,” serve as reference identities throughout our study.

\paragraph{4D Reconstruction as a Data Source.}
To obtain sufficient customization data with multi-view identity information and camera information, we propose a novel approach: re-purposing 4D Gaussian Splatting (4DGS) based on ~\citep{yang2023gs4d, 4drotor}, originally designed for 4D reconstruction, as a data generator for video generation tasks. While 4DGS excels at producing high-fidelity multi-view renderings along diverse camera trajectories, it lacks the capability to synthesize novel content beyond the captured scenes. Conversely, video generation models can create new content and contexts, but often struggle with maintaining precise subject identity and view-dependent consistency. By leveraging 4DGS-generated data to customize the video generation model, we combine the strengths of both paradigms: precise subject modeling from 4DGS and creative generalization from video generation. Specifically, we reconstruct each captured sequence using 4DGS and render videos along diverse camera trajectories, generated by randomly sampling starting and ending positions within a 2–10 meter radius and linearly interpolating between them to create smooth motion paths. 
To further enrich the data with lighting diversity, we apply a generalizable video relighting model~\cite{mei2025lux} using HDRI maps from Poly Haven~\cite{PolyHavenHDRIs}.

\paragraph{CG and Real-Life Videos.}
To support application in subject–scene composition and evaluate our method in both controlled and real-world settings, we incorporate two additional data sources: CG-rendered scenes and real-life videos. For CG data, 
we use Blender Cycles to render photorealistic 3D indoor environments with diverse camera trajectories, providing precise camera annotations for training scene-customized models. For real-life data, we capture handheld videos of scenes and dynamic subjects using an iPhone, introducing natural camera motion. We estimate camera poses and intrinsics using CUT3R~\cite{wang2025continuous}, enabling subject and scene customization from accessible, real-life video. These sources enable our framework to generalize beyond studio captures, supporting novel camera paths and real-world contexts.

\subsection{Multi-subject generation}

\paragraph{Joint Training.}
To enable the generation of multiple entities—such as two subjects or a subject and a scene—within the same video under novel contexts, we adopt a joint training strategy. Specifically, we fine-tune the model on separate single-entity customization datasets, where each video contains either a single subject or a single scene. Despite training on disjoint examples, the model learns to compose multiple customized entities during inference. For multi-subject generation, we further incorporate joint-subject data—videos featuring both subjects together—to improve the realism and coherence of inter-subject interactions. Each customized entity is associated with a unique token during training, and these tokens are combined at inference time to generate videos featuring multiple customized components.

\paragraph{Customization via Independently Customized Models with Noise Blending.}
We propose an alternative to joint training for multi-subject video generation by composing independently customized models at inference time, avoiding the need to retrain for every subject combination.
Inspired by OMG~\citep{kong2024omg}, originally designed for multi-subject image generation, our method leverages independently customized text-to-video (T2V) models—each fine-tuned separately for a specific subject—and integrates their outputs through a noise-blending strategy. The method comprises two stages:
The first step is focused on \textbf{Spatio-Temporal Layout}.
We first generate a coarse layout video without identity-specific customization, using a generic prompt (e.g., ``a man and a woman in a coffee shop'') that excludes subject-specific tokens. This video establishes plausible spatial and temporal arrangements of subjects. We then use SA2VA~\citep{yuan2025sa2va}, based on SAM2~\citep{ravi2024sam}, to segment each subject, producing spatio-temporal masks $M_i$. These masks are extended across the entire video by assigning each pixel to the nearest segmented region, ensuring complete spatial-temporal coverage.
The second step is \textbf{personalized generation via Noise Blending.}
To retain layout while incorporating subject-specific identities, we adopt a two-phase denoising procedure. Beginning from the same initial noise seed as Step 1, we first perform the initial 10\% of denoising steps without customization to preserve the coarse spatial-temporal layout. This threshold provides a balance: setting it too high reduces identity fidelity, while setting it too low destabilizes the scene layout. Subsequently, at each remaining timestep $t$, we predict the next latent state $z_{t-1}^{i}$ separately for each subject $i$ using its corresponding customized model $\mathrm{T2V}^{i}$: $z_{t-1}^i = \mathrm{T2V}^i(z_t, p^i, t)$, where $p^{i}$ is a modified version of the original prompt in which only the target subject is replaced by its specific identity token. We then blend these predictions according to the segmentation masks: $z_{t-1} = \sum_{i} M_i * z_{t-1}^{i}$. This approach accurately customizes each subject region while ensuring coherent global spatio-temporal consistency.
\section{Experiments}
\label{sec:experiments}

\subsection{Datasets}

\begin{figure*}[t!]
    \centering 
    \includegraphics[width=0.94\textwidth]{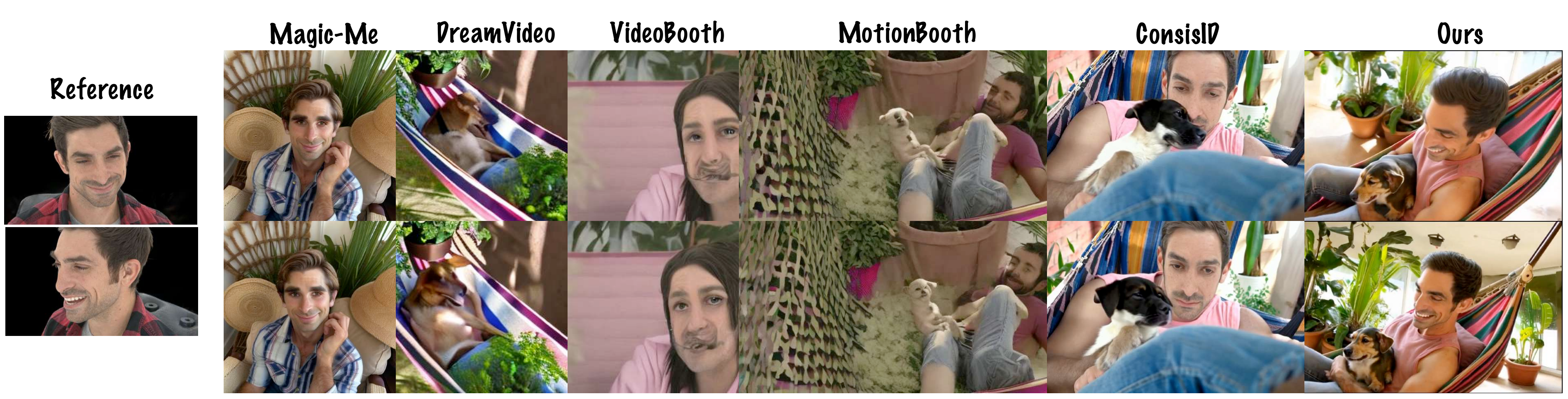}
    \caption{Customization results of T2V baselines and our method, demonstrating superior multi-view identity preservation by our approach.}
    \label{fig:ID_baselines}
\end{figure*}

\paragraph{4DGS from Volumetric Captures.}
To validate our method in a controlled studio setting, we capture facial and full-body performances of two subjects, “Alex” and “Emily.” Each sequence is reconstructed using 4DGS~\citep{yang2023gs4d}, and rendered with randomly generated moving camera trajectories to create diverse multi-view videos. In total, each subject has 256 videos across 8 performance sequences. To introduce lighting diversity, we apply a generalizable video relighting model~\citep{mei2025lux} using HDRI maps from Poly Haven~\citep{PolyHavenHDRIs}, generating an additional 128 relit videos per subject. For multi-subject generation, we render 27 joint-subject videos featuring both subjects across 3 sequences.

\paragraph{CG Scenes.}
We construct a dataset of 10 artist-designed indoor scenes rendered with Blender Cycles, each featuring 24 diverse camera trajectories. After filtering out invalid paths, 16 valid trajectories per scene are retained for subject–scene composition. Each rendered video includes ground-truth camera annotations.

\paragraph{Real-Life Videos.}
We capture two indoor environments and one dynamic subject (a cat) using handheld iPhone videos, each about one minute long and recorded while walking to introduce natural camera motion and viewpoint diversity. Each video is split into 20 two-second clips, with camera poses and intrinsics estimated using CUT3R~\cite{wang2025continuous}. This setup supports effective subject and scene customization under real-world conditions.

\subsection{Baselines} 

\paragraph{Camera Control.}
We select 2 contemporary camera control models: AC3D~\citep{bahmani2025ac3d} and CameraCtrl~\citep{he2024cameractrl}. We choose AC3D's implementation based on CogVideoX~\cite{yang2024cogvideox}, a diffusion transformer based model, whereas CameraCtrl uses a UNet-based video diffusion backbone~\citep{guo2024animatediff}.

\paragraph{Personalization.}
To validate our method can generate character in consistent with the identity source, we compare our method with ConsisID~\cite{huang2024consistentid}, Videobooth~\cite{jiang2024videobooth}, Magic-Me~\cite{ma2024magic}, Dreamvideo~\cite{wei2024dreamvideo} and Motionbooth~\cite{wu2024motionbooth}. Notably, MotionBooth also enables camera control during personalization. However, its camera motion is limited to 2D movements, such as panning, which restricts its applicability. In contrast, our method can generate 3D camera motions for personalized subjects, offering greater flexibility and realism.

\subsection{Evaluation metrics} 

\paragraph{Multi-view Identity Preservation.}
We collect 10 reference images per subject, capturing a variety of viewpoints, including frontal and profile views.
For all methods, we first generate videos using the same set of 100 text prompts.
Next, on the generated frames, we run SCRFD face detection~\cite{guo2021sample} and compare the similarity between the cropped face and the reference faces using AdaFace~\cite{kim2022adaface}. We then take the maximum similarity score across multiple reference faces from the same identity source. We discard frames where no face is detected. 

\paragraph{Camera Control.}
We use rotation and normalized translation errors~\citep{he2024cameractrl} estimated by CUT3R~\citep{wang2025continuous} assess camera steerability.
We evaluate all methods on the same 200 test text prompts and camera trajectories from the RealEstate10K dataset.

\paragraph{General Video Quality.}
We assess text controllability by computing the average CLIP~\cite{radford2021learning} similarity between the prompt and generated frames. Temporal consistency is measured via average CLIP image similarity between consecutive frames. 
Additionally, we benchmark our method using four metrics from VBench~\cite{huang2024vbench}, specifically evaluating subject consistency, background consistency, and temporal flickering to provide a comprehensive assessment of video quality.


\subsection{Identity preservation} \label{sec: ID_preservation_T2V}

\paragraph{Comparison with Baselines. }
The quantitative comparison of identity preservation and general video generation quality is shown in~\cref{tab:face_evaluation}, with illustrative examples from each method presented in~\cref{fig:ID_baselines}. Notably, for multi-view identity preservation, our model achieves the highest AdaFace score among all baselines, surpassing ConsisID, which relies solely on a single facial image during inference and consequently struggles to maintain consistent identities across multiple views. This underscores the importance of utilizing multi-view datasets, as employed by our method, for enhancing identity consistency in personalized multi-view video generation. Although the fine-tuning-based MagicMe method achieves superior scores in terms of subject consistency, background consistency, and reduced temporal flickering, we observe that it suffers from substantial identity degradation and generates notably less motion compared to our approach.
Also, MotionBooth, which enables 2D camera control, fails to generate identity-preserving videos.

\begin{table*}[htp!]
    \centering
    \caption{Quantitative results for customization. User study columns report the percentage of responses favoring each method for each evaluation criterion. $\Uparrow$/$\Downarrow$ indicates a higher/lower value is better. \textbf{Bold} indicates the best results. \colorbox{lightgray}{Gray-shaded cells} indicate values that are significantly lower than the others.}
    
    \resizebox{0.9\linewidth}{!}{
    {
    \begin{tabular}{l|c|c|c|c|c|c|c|c|c|c}
        \toprule

        & & & & \multicolumn{4}{c|}{VBench $\Uparrow$} & \multicolumn{3}{c}{UserStudy $\Uparrow$} \\
        
         & \makecell{AdaFace\\$\Uparrow$}
         & \makecell{CLIP-T\\$\Uparrow$}
         & \makecell{CLIP-I\\$\Uparrow$}
         & \makecell{Subject\\Consistency}
         & \makecell{Background\\Consistency}
         & \makecell{Temporal\\Flickering}
         &
         \makecell{Dynamic\\Degree}
         &
         \makecell{Multi-view \\ Identity}
         & 
         \makecell{Facial \\ Realism}
         & 
         \makecell{Text \\ Alignment}
         \\
        \midrule
        & \multicolumn{9}{c}{\textbf{Text-to-video customization}}\\
        MagicMe &   0.280     &   0.303 &	\textbf{0.991} &	\textbf{0.978} &	\textbf{0.967} &	\textbf{0.984} & \cellcolor{lightgray}0.15  & 3.18\% & 10.13\% & 1.85\%            \\
        DreamVideo & 0.194 &   0.318 &	0.961 &	0.893 &	0.918 &	0.958 & 0.44 & 0.98\% & 0.99\% & 1.16\% \\
        VideoBooth &    0.279   &  0.274 &	0.966 &	0.909 &	0.941 &	0.967 & 0.55  & 1.54\% & 0.99\% & 1.62\%         \\
        MotionBooth &    \cellcolor{lightgray}0.191     &    0.324    &   0.954     &      0.905               &        0.904                &       0.927   & \textbf{1.0} & 2.20\% & 0.66\% & 1.16\%      \\
        ConsisID    &    0.301     &  0.355      &    0.981    &   0.882                  &      0.892                  &       0.963 & 0.36 & 12.96\% & 17.30\% & 21.29\%             \\
        Ours        &   \textbf{0.351}      &    \textbf{0.356}    &  \textbf{0.991}      &       0.933              &    0.946                    &      0.975     & 0.72 & \textbf{81.34\%} & \textbf{70.59\%} & \textbf{74.07\%}        \\
        Ours (frontal-only)       &   0.327     &    0.353    &  0.989      &       0.929              &    0.952                    &      0.980      & 0.59 & - & - & -         \\
        \midrule
        & \multicolumn{9}{c}{\textbf{Image-to-video customization}} \\
        Non-customized & 0.324 & 0.343 & 0.980 & 0.898 & 0.925 & 0.953 & 0.78 & 34.57\% & - & - \\
        Ours-I2V & \textbf{0.350}  & \textbf{0.345} & \textbf{0.984} & \textbf{0.908} &	\textbf{0.932} & \textbf{0.960} & 0.72 & \textbf{65.43\%} & - & -  \\
        \bottomrule
    \end{tabular}%
    }
    }
    \label{tab:face_evaluation}
\end{table*}

\paragraph{User Study on Baseline Comparison. }
We conducted a user study with 19 participants to compare video generation methods on multi-view identity preservation, facial realism, and text alignment across 60 prompts featuring two reference identities (Emily or Alex). Participants viewed multi-view reference images of each identity and selected the best video per criterion. Our method was preferred in 81.3\% of cases for identity preservation, 70.6\% for facial realism, and 74.1\% for text alignment.

\begin{figure}[htp!]
    \centering 
    \includegraphics[width=0.9\columnwidth]{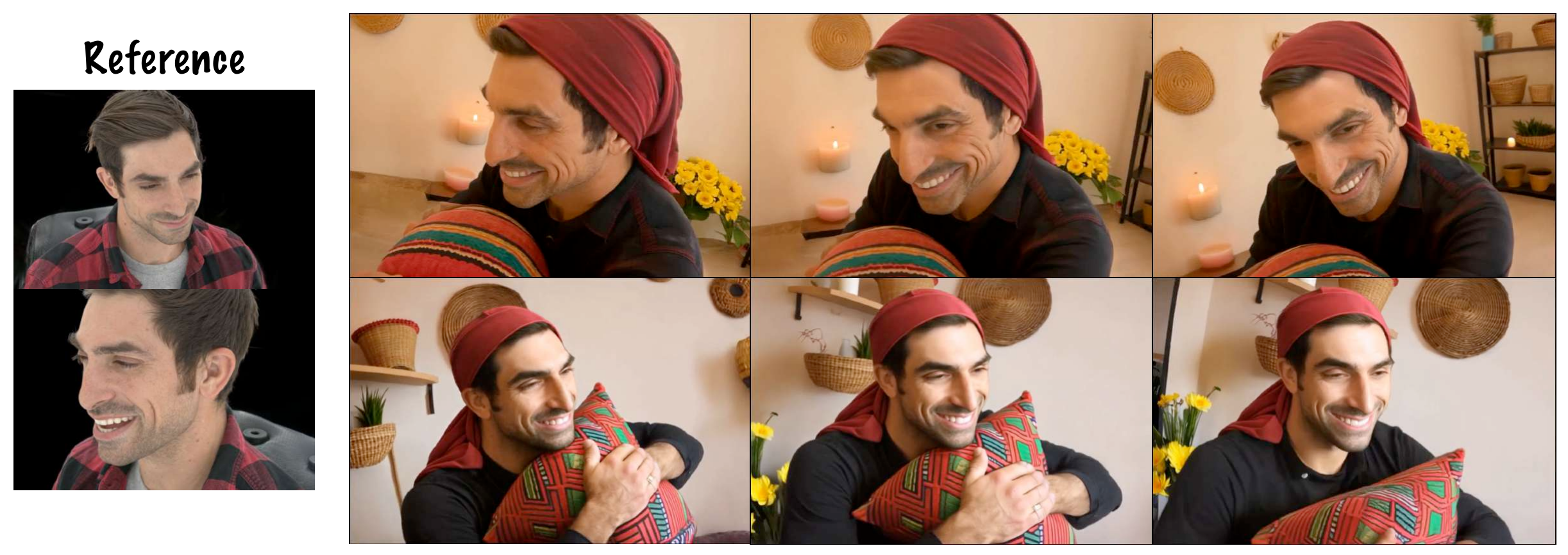}
    \caption{Generated videos with multi-view data (top) and frontal-view-only data (bottom). Multi-view training yields markedly better identity preservation across viewpoints.
    }
    \label{fig:multiview}
\end{figure}

\paragraph{Effects of Multi-view Data.}
To further investigate the impact of multi-view training data, we performed an ablation study using only frontal-view training images. Quantitative results are presented in~\cref{tab:face_evaluation}, showing a noticeable decrease in AdaFace scores compared to models trained on the complete multi-view dataset. Qualitative examples (\cref{fig:multiview}) also illustrate poorer identity preservation from side-view angles when trained exclusively on frontal-view data. These results underscore the importance of multi-view data for effective multi-view identity preservation.

\begin{figure} [htp!]
    \centering 
    \includegraphics[width=0.9 \columnwidth]{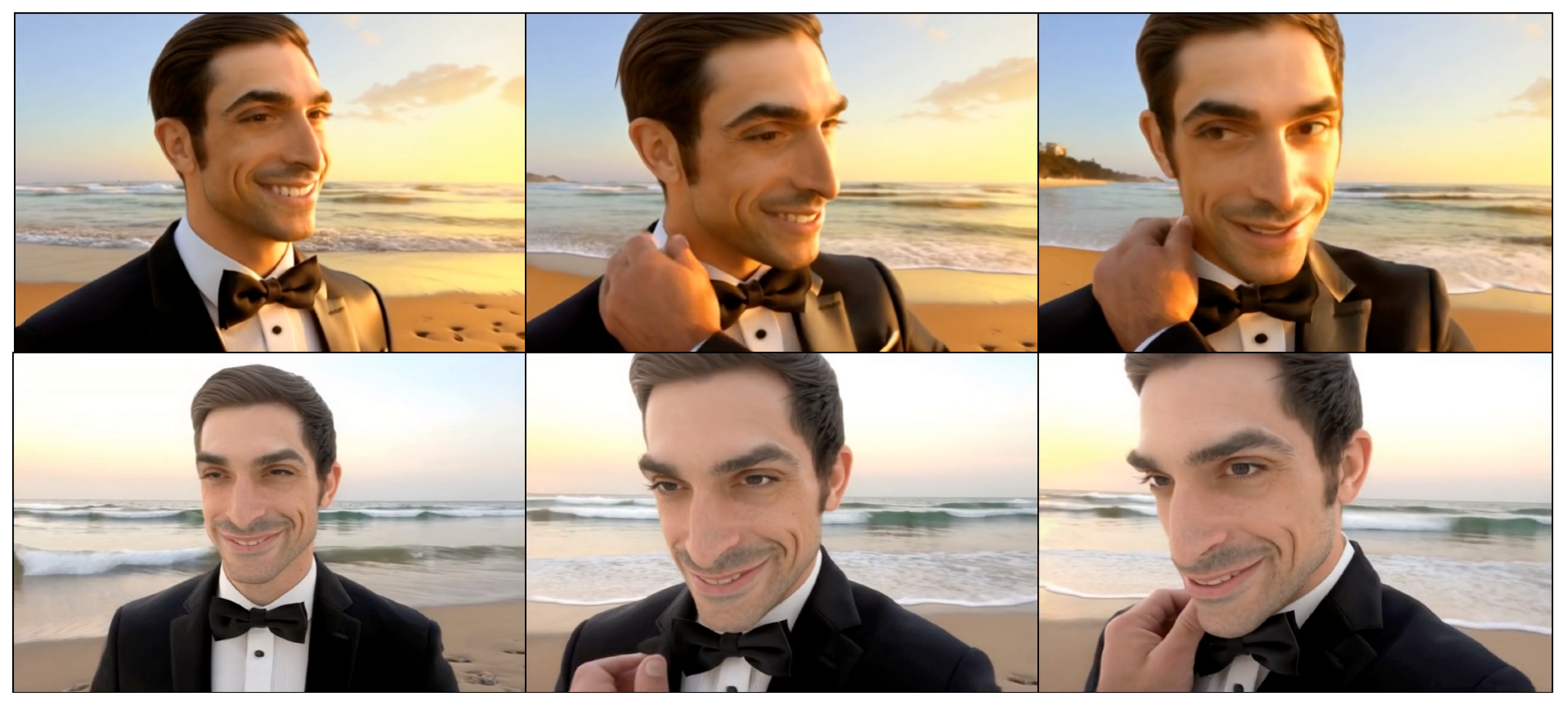}
    \caption{Generated videos with additional relit data (top) and without (bottom). Relit data significantly enhances lighting realism and diversity.
    }
    \label{fig:lighting}
\end{figure}

\paragraph{Effects of Relit Data.}
To evaluate the impact of relit 4DGS data on lighting realism, we conducted an ablation study comparing models trained with and without it. In a user study across 60 prompts, 18 participants preferred the relit-data model in 83.9\% of cases. As shown in~\cref{fig:lighting}, relit data significantly enhances lighting realism, while its absence results in flatter illumination.

\subsection{Camera control}

A quantitative comparison of camera controllability is presented in~\cref{tab: camera_metrics}. Our pretrained camera-conditioned model achieves the lowest translation and rotation errors, establishing a strong foundation for further camera-conditioned customization.

\begin{figure*}[htp!]
    \centering 
    \includegraphics[width=\textwidth]{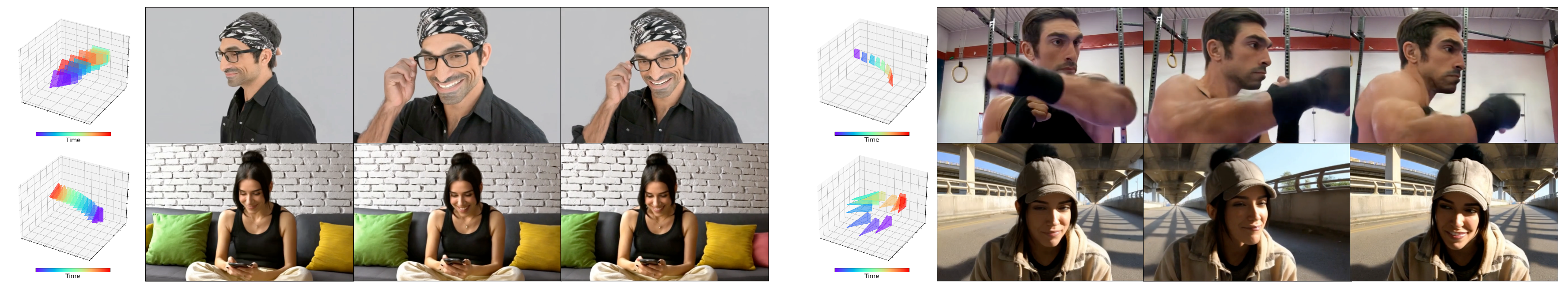}
    \caption{Text-to-video (T2V) generation results with camera control for a single subject.
    }
    \label{fig:T2V_camera}
\end{figure*}

\paragraph{Multi-view Customization with Camera Control. }
\Cref{fig:T2V_camera} shows qualitative examples of videos generated after subject-specific customization with camera control, along with the corresponding input camera trajectories. Our method faithfully follows the input camera path, generating temporally coherent videos where the subject’s appearance is consistently maintained across different viewpoints. It also handles background changes smoothly and ensures that the subject, background, and lighting evolve together in a visually plausible manner as the camera moves. This coherence is supported by our multi-view-aware customization training, which enables the model to maintain consistent subject identity and spatial relationships under dynamic camera motion.

\paragraph{Effects of Moving Camera during Customization. }
To study the effect of camera motion during customization, we fine-tune the model on two dataset variants: one with dynamic trajectories (Ours-customized) and one with static cameras (Ours-customized-static). As shown in~\cref{tab: camera_metrics}, removing camera motion increases translation and rotation errors, highlighting its importance for maintaining camera controllability.

\begin{table}[ht]
\centering
\caption{Quantitative comparisons for camera control. $\Uparrow$/$\Downarrow$ indicates a higher/lower value is better. \textbf{Bold} indicates the best results.}
\resizebox{\linewidth}{!}{
\begin{tabular}{lcccc}
\toprule
 & TransErr $\Downarrow$ & RotErr $\Downarrow$ & CLIP-T $\Uparrow$ & CLIP-I $\Uparrow$ \\
\midrule
CameraCtrl &  0.522 & 0.163 & 0.301 & 0.965  \\
AC3D       &  0.310 & 0.112 & 0.329 & 0.990\\
Ours       &  \textbf{0.267} & \textbf{0.047} & \textbf{0.332} & \textbf{0.991}  \\
\midrule
Ours-customized & 0.324 & 0.086 & 0.321 & 0.993 \\
Ours-customized-static & 0.482 & 0.125 & 0.330 & 0.991 \\
\bottomrule
\label{tab: camera_metrics}
\end{tabular}
}
\end{table}

\begin{figure*}[htp!]
    \centering 
    \includegraphics[width=\textwidth]{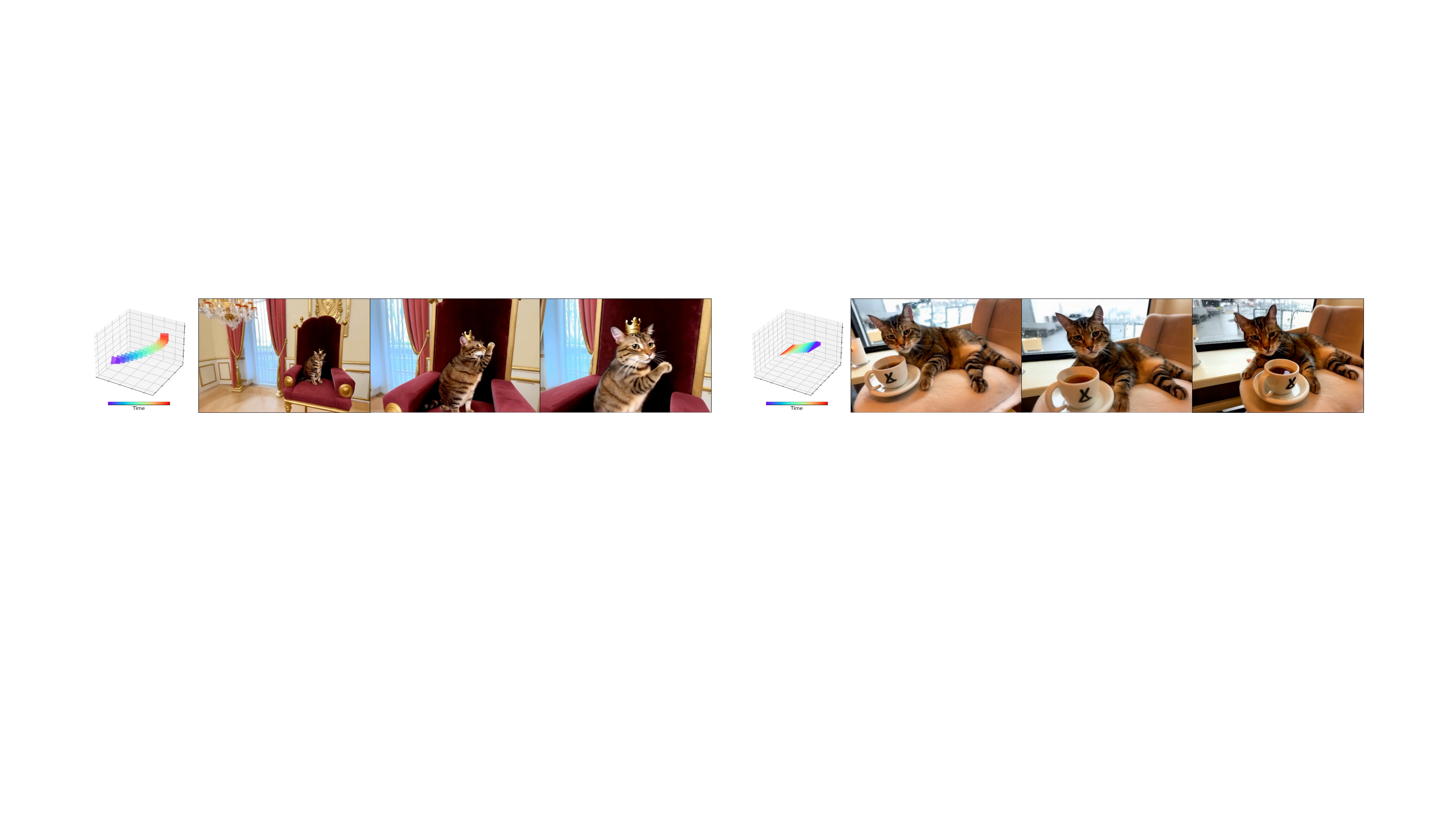}
    \caption{Customization results from real-life videos of a cat.
    }
    \label{fig:cat}
\end{figure*}

\paragraph{Real-life Videos.}
Beyond 4DGS human data, we demonstrate the effectiveness of our method on a real-world customization dataset. \Cref{fig:cat} showcases a dataset featuring a cat captured under diverse camera angles, viewpoints, poses, and environments, along with generated videos of the same cat in novel contexts. The results show that our method successfully preserves the cat’s identity across views and supports controllable camera motion, highlighting its applicability to in-the-wild scenarios.

\subsection{Multi-subject generation} \label{sec: multi_subject_generation_T2V}

\begin{figure*}[htp!]
    \centering 
    \includegraphics[width=\textwidth]{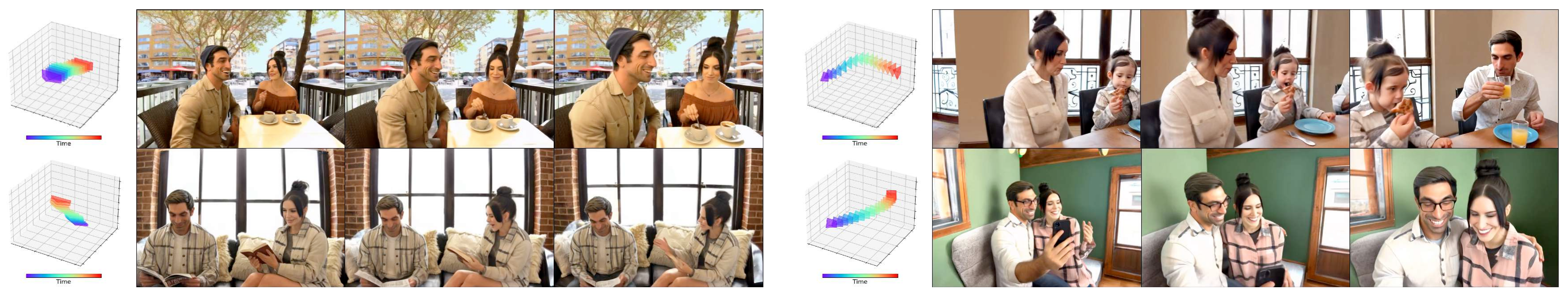}
    \caption{Text-to-video (T2V) generation results with camera control for multiple subjects.
    }
    \label{fig:T2V_multi_subject}
\end{figure*}

\paragraph{Multi-subject Interaction.}
\Cref{fig:T2V_multi_subject} presents qualitative examples of videos generated by our customized model with camera control, which was trained on separate single-subject customization datasets (each video containing either subject) as well as a small joint-subject dataset featuring both subjects together. The results demonstrate that our model can accurately generate both subjects in the same scene, maintaining strong multi-view identity consistency for each individual. Moreover, the interactions appear natural and coherent, showing the model captures both individual traits and their spatial and behavioral relationships.

\begin{figure}[htp!]
    \centering 
    \includegraphics[width=0.9\columnwidth]{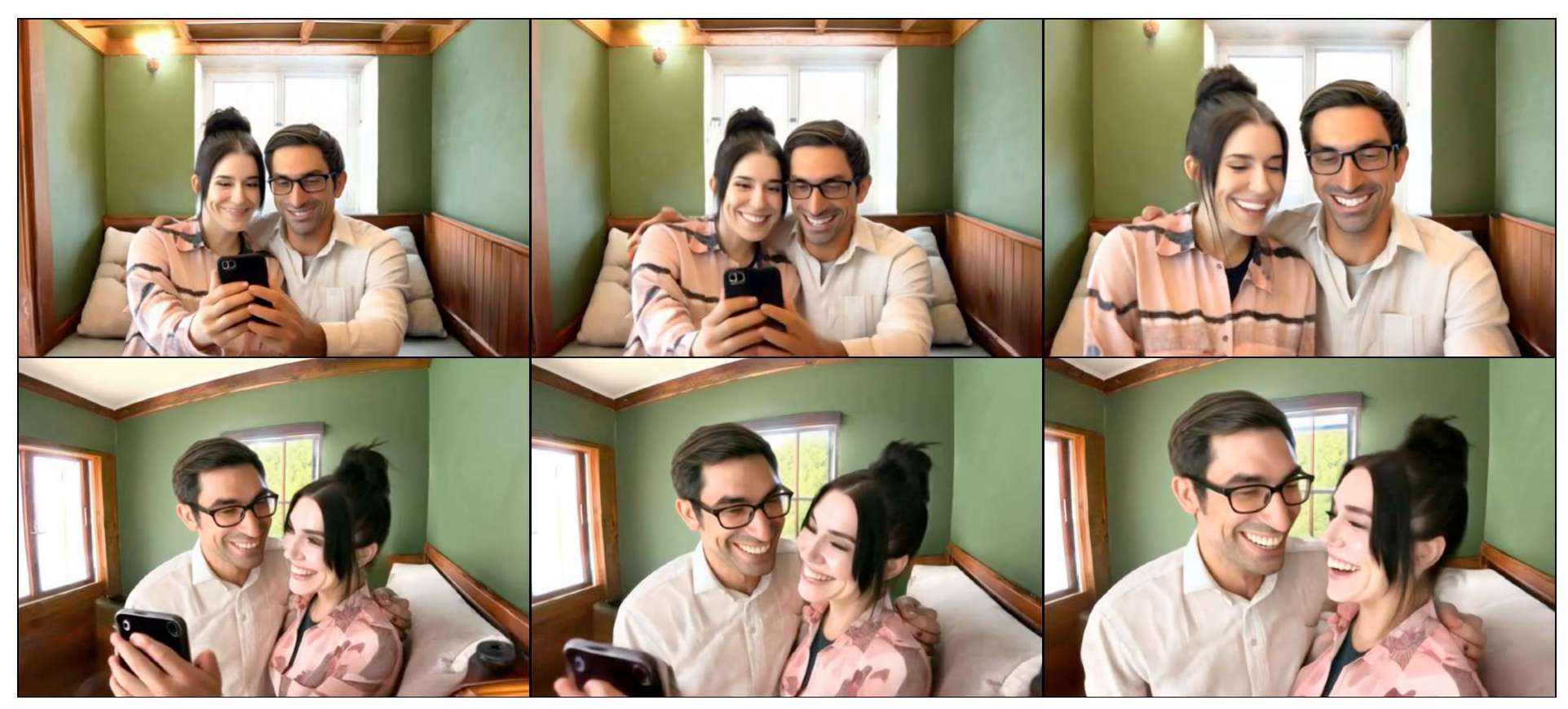}
    \caption{Generated videos with (top) and without (bottom) joint-subject data. Including joint-subject data improves multi-subject interaction quality.
    }
    \label{fig:vps07}
\end{figure}

\paragraph{Effect of joint-subject data. }
To assess the impact of joint-subject data—videos featuring both subjects—we conducted an ablation study comparing models trained with and without it. In the ablated setting, the model was trained only on single-subject datasets, where each video contained either subject but not both. As shown in~\cref{fig:vps07}, adding joint-subject data improves spatial relationships and interaction realism. A user study across 60 prompts with 18 participants further confirmed this: 72.9\% preferred videos from the model trained with joint-subject data, underscoring its importance for realistic multi-subject interaction.

\begin{figure*}[htp!]
    \centering 
    \includegraphics[width=\textwidth]{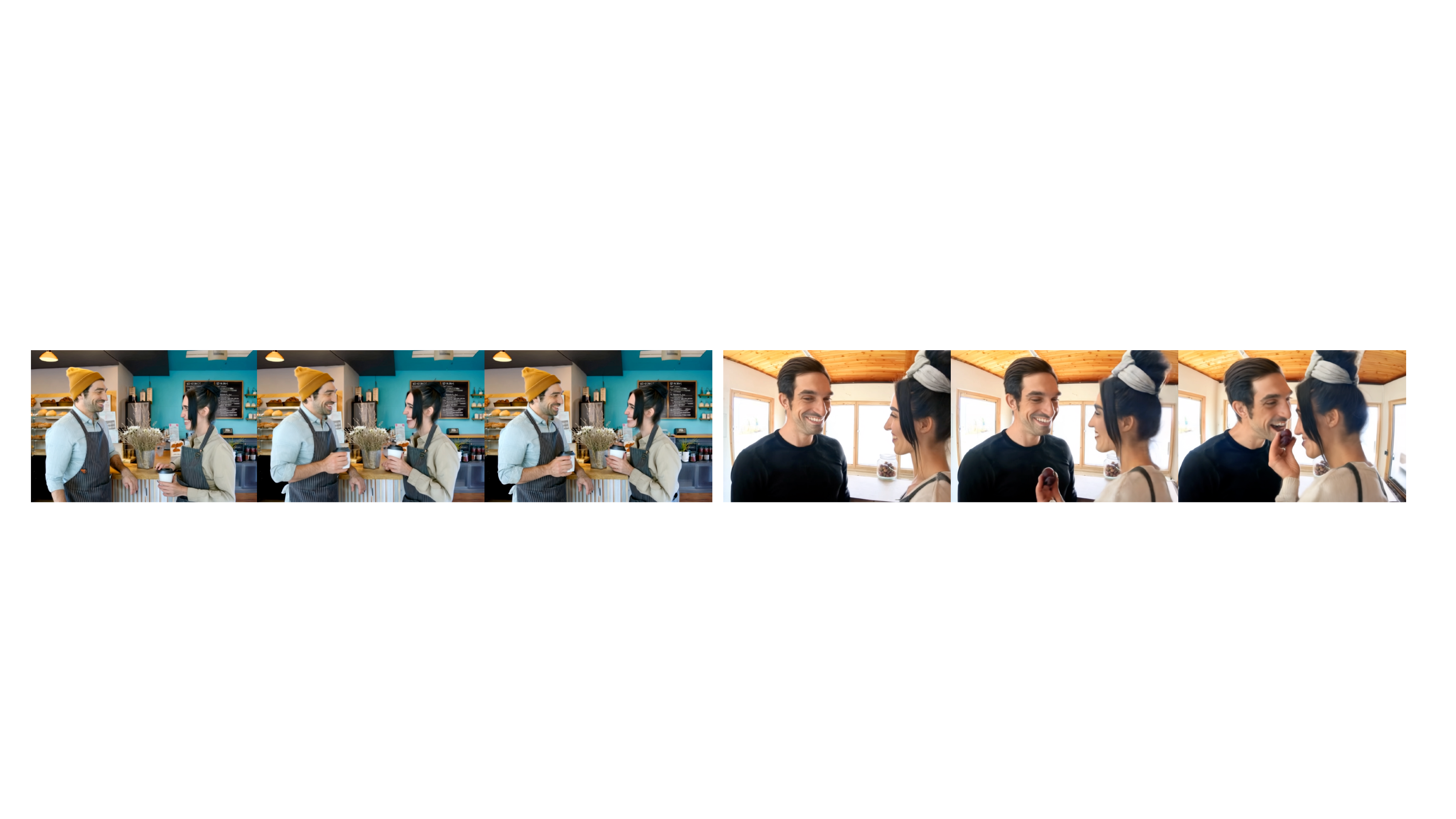}
    \caption{Generation results using noise blending with independently customized models for multi-subject video synthesis.
    }
    \label{fig:T2V_multi_subject_OMG}
\end{figure*}

\paragraph{Noise Blending Enables Synergy between Independently Customized Models. }
We use a noise blending technique to combine independently customized models at inference time for multi-subject video generation.
As shown in~\cref{fig:T2V_multi_subject_OMG}, this approach successfully generates scenes featuring both subjects and captures plausible interactions between them—even though no model was trained on data containing both individuals. 
The noise blending technique achieves an AdaFace score of 0.320, slightly lower than the 0.337 obtained with joint training. However, unlike joint training, it enables flexible and modular multi-subject generation without requiring retraining on combined datasets.

\subsection{Image-to-video customization} \label{sec: exp_I2V}

\begin{figure}[htp!]
    \centering 
    \includegraphics[width=0.95\columnwidth]{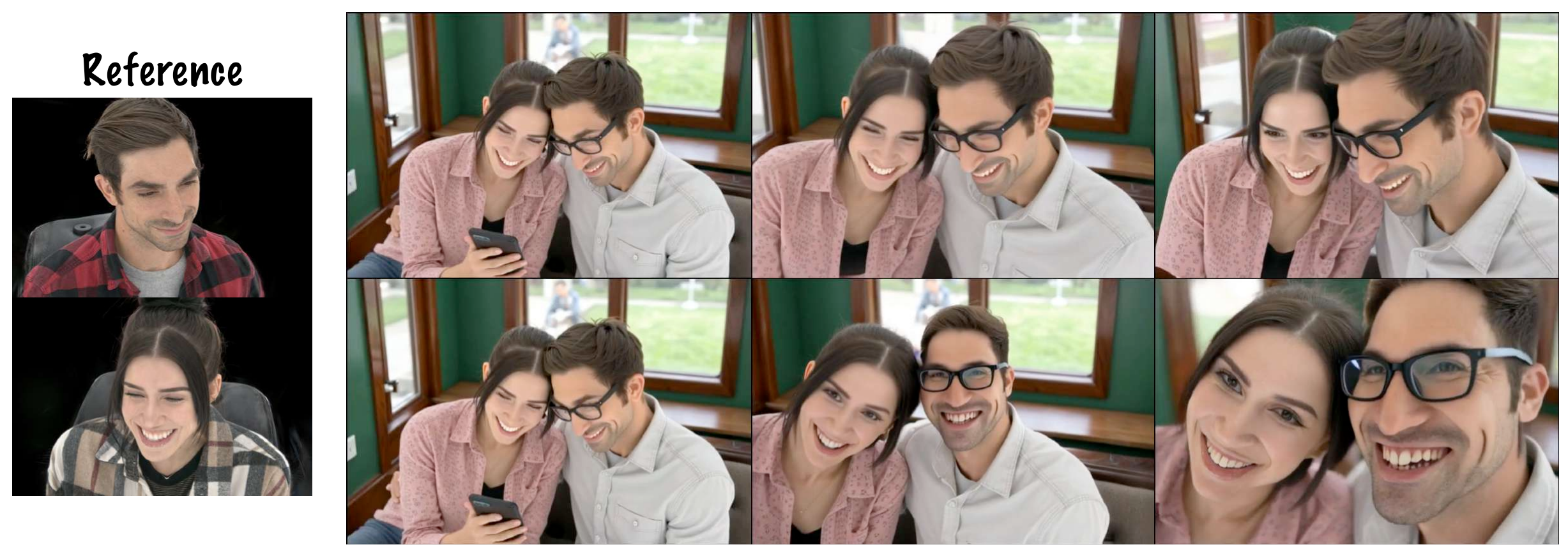}
    \caption{Generated videos with (top) and without (bottom) image-to-video customization. Customization improves multi-view identity preservation.
    }
    \label{fig:I2V_pretrained_customized}
\end{figure}

\begin{figure*}[htp!]
    \centering 
    \includegraphics[width=\textwidth]{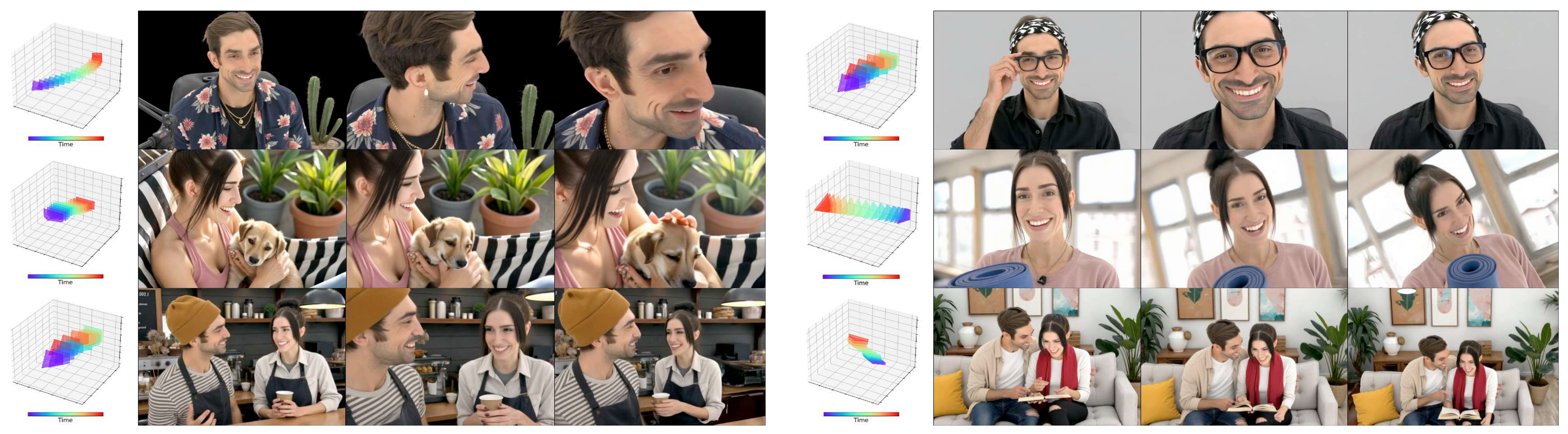}
    \caption{Image-to-video (I2V) generation results with camera control for single and multiple subjects.
    }
    \label{fig:I2V_camera}
\end{figure*}

\paragraph{Effect of Image-to-Video Customization.}
As shown in~\cref{fig:I2V_camera}, the customized I2V model generates identity-consistent videos for both single and multiple subjects while accurately following camera trajectories. To assess the need for I2V customization—even when the initial frame from the T2I model is accurate—we compare a customized I2V model fine-tuned on subject-specific multi-view data with a pre-trained, non-customized model.
(1) Qualitative results in~\cref{fig:I2V_pretrained_customized} show that the non-customized model exhibits significant identity drift, while the customized model maintains subject’s identity.
(2) Quantitative results in~\cref{tab:face_evaluation} show higher AdaFace scores and improved subject consistency, background consistency, and temporal stability, confirming that I2V customization yields more robust results for personalized subjects.
(3) In a user study with 18 participants across 60 prompts, 65.4\% preferred the customized model in terms of identity preservation.
Together, these results highlight the importance of customizing the I2V model for achieving high-quality, identity-consistent video generation.

\subsection{Other applications}

\begin{figure*}[htp!]
    \centering 
    \includegraphics[width=\textwidth]{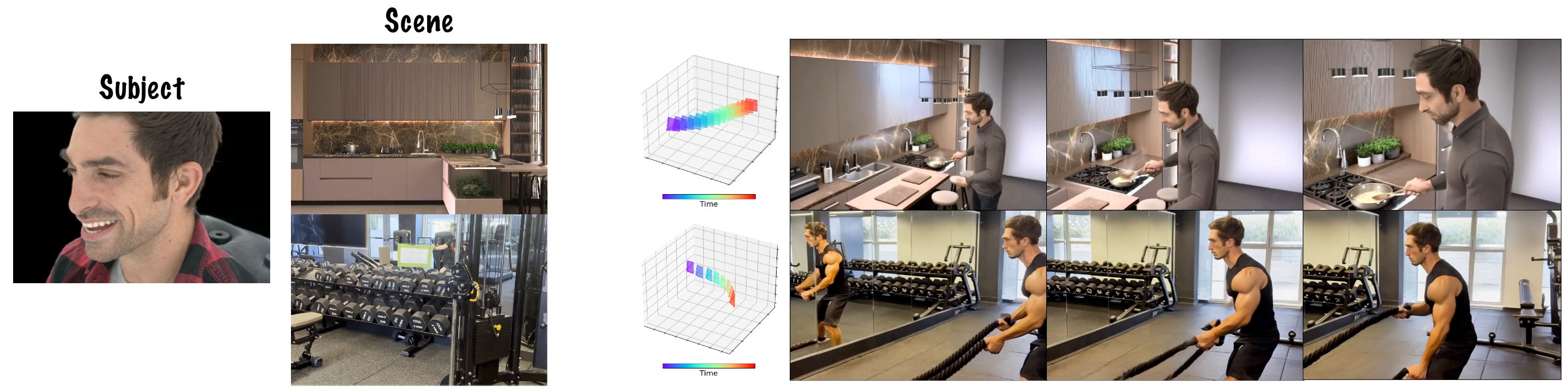}
    \caption{Customization of a single subject across two scenes with camera control, demonstrating subject–scene interaction during generation.
    }
    \label{fig:T2V_scene_singleSubject}
\end{figure*}

\begin{figure*}[htp!]
    \centering 
    \includegraphics[width=\textwidth]{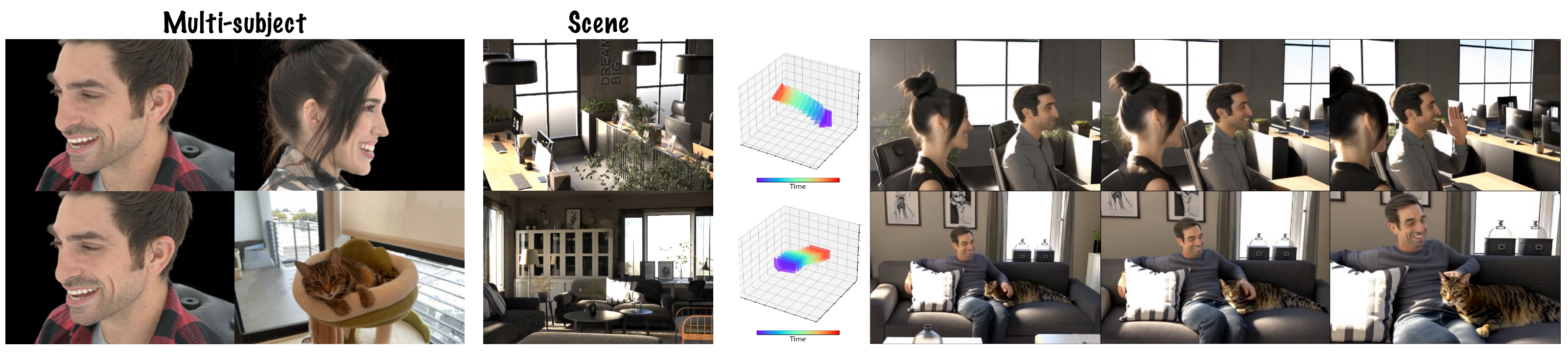}
    \caption{Customization of multiple subjects across two scenes with camera control, demonstrating both subject–scene and inter-subject interactions during generation.
    }
    \label{fig:T2V_scene_multiSubject}
\end{figure*}

\paragraph{Scene Customization.}
Beyond subject customization, our method supports scene customization to enable subject–environment interactions. We fine-tune on scene-specific datasets with varying camera trajectories. \Cref{fig:T2V_scene_singleSubject} shows a single-subject case with natural scene interaction, while \Cref{fig:T2V_scene_multiSubject} illustrates realistic multi-subject interactions within the customized scene.

\begin{figure*}[htp!]
    \centering 
    \includegraphics[width=\textwidth]{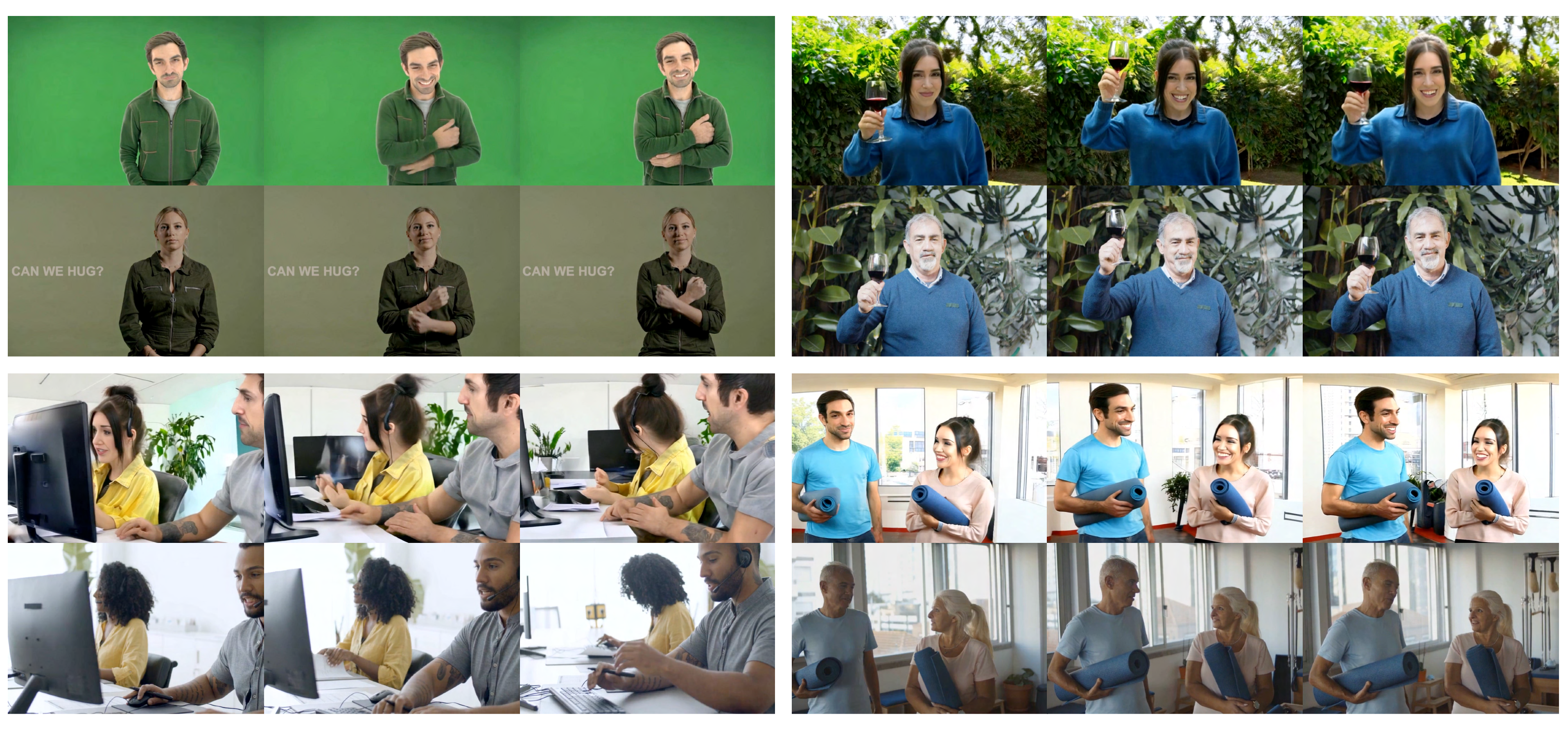}
    \caption{Customization with motion and layout control. The first and third rows show generated videos; the second and fourth rows show the corresponding source videos. The results demonstrate that the generated subjects preserve both spatial layout and motion from the source videos.
    }
    \label{fig:GWTF}
\end{figure*}

\paragraph{Customization with Motion and Layout Control.}
Beyond camera control, we enable control over subject motion and spatial layout by fine-tuning the Go-with-the-Flow T2V model~\citep{burgert2025go} on our customization dataset. This model uses optical flow from a source video as a control signal, allowing synchronized control over camera and object movement. Given a source video with humans, we aim to preserve its motion and layout while replacing the subjects with customized ones. We extract optical flow and pair it with the source prompt as input to the customized model. As shown in~\cref{fig:GWTF}, the generated videos largely maintain the original motion and layout while generating the target subject’s appearance.

\section{Conclusion}
\label{sec:conclusion}

We have introduced a framework that addresses two key challenges in video generation for filmmaking: customization for multi-view identity preservation and precise camera control. Central to our approach is a novel customization data pipeline that combines volumetric capture, 4DGS-based re-rendering for diverse, accurately annotated camera trajectories, and relightable augmentation. Fine-tuning on this dataset improves identity fidelity, camera conditioning, and lighting adaptability. Additionally, our framework supports core virtual production capabilities, including multi-subject generation via joint training and noise blending, scene and real-life customization, and motion-aware spatial layout control. Together, these components offer a scalable and flexible solution for controllable, high-fidelity video generation in virtual production.
Our limitations include the need for fine-tuning to fully leverage high-quality multi-view 4DGS data and the low resolution of the CogVideoX backbone, which underutilizes these higher-resolution inputs.

\begin{acks}
We would like to express our gratitude to Stephan Trojansky and Jeffrey Shapiro for their initial and ongoing executive support; Sebastian Sylwan, Daniel Heckenberg, Jitendra Agarwal, Matheus Leao, and Sungmin Lee for their IT support; Xueming Yu and David George for their hardware support; Jennifer Lao and Lianette Alnaber for their operational support; and Winnie Lin, Lukas Lepicovsky, Ashish Rastogi, Ritwik Kumar, Cornelia Carapcea, and Girish Balakrishnan for their insightful technical discussions.
\end{acks}

\clearpage

\clearpage
\bibliographystyle{ACM-Reference-Format}
\bibliography{references}

\clearpage
\appendix












\section{Appedix} 
\label{sec:supp_user_study}

We conducted four user studies, detailed below.

\subsection{User study on identity preservation compared to baselines} \label{app: user_study_comparison_with_baselines}
We conducted a user study with 19 participants to evaluate six different video generation methods in terms of multi-view identity preservation, facial realism, and text-video alignment. Participants assessed videos generated by both baseline methods and our approach across 60 prompts, each associated with one of two reference identities (Emily or Alex). 

For each prompt, participants were provided with multi-view reference images of the target identity and asked to select the best-performing video for each evaluation criterion. The evaluation questions were as follows:
\begin{itemize}
    \item \textbf{Multi-view identity preservation:} ``Which video best preserves the identity of the subject in the reference images?''
    \item \textbf{Facial realism:} ``Which video shows the most natural and realistic integration of the human face (i.e., the face does not appear artificially pasted onto the scene)?''
    \item \textbf{Text-video alignment:} ``Which video aligns most accurately with the given text prompt in terms of content, actions, and overall depiction?''
\end{itemize}

\begin{figure*}[htp]
  \centering

  \begin{subfigure}{\linewidth}
    \centering
    \includegraphics[width=\linewidth]{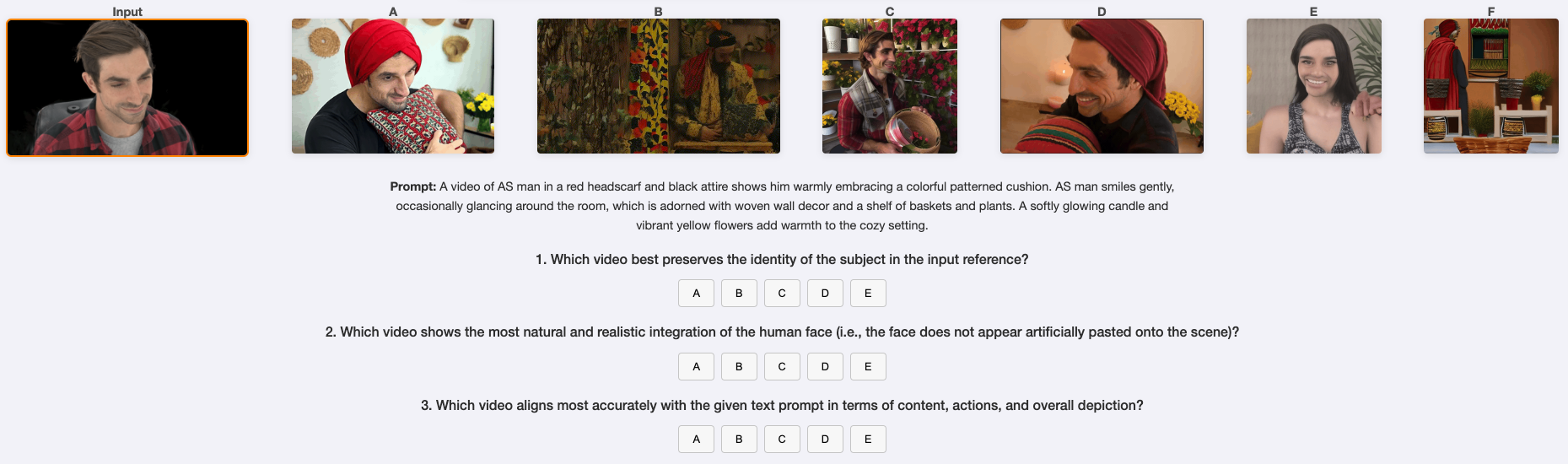}
    \caption{User-study interface and questions for comparison with baselines.}
  \end{subfigure}
  \vspace{6pt} 

  \begin{subfigure}{0.32\linewidth}
    \centering
    \includegraphics[width=\linewidth]{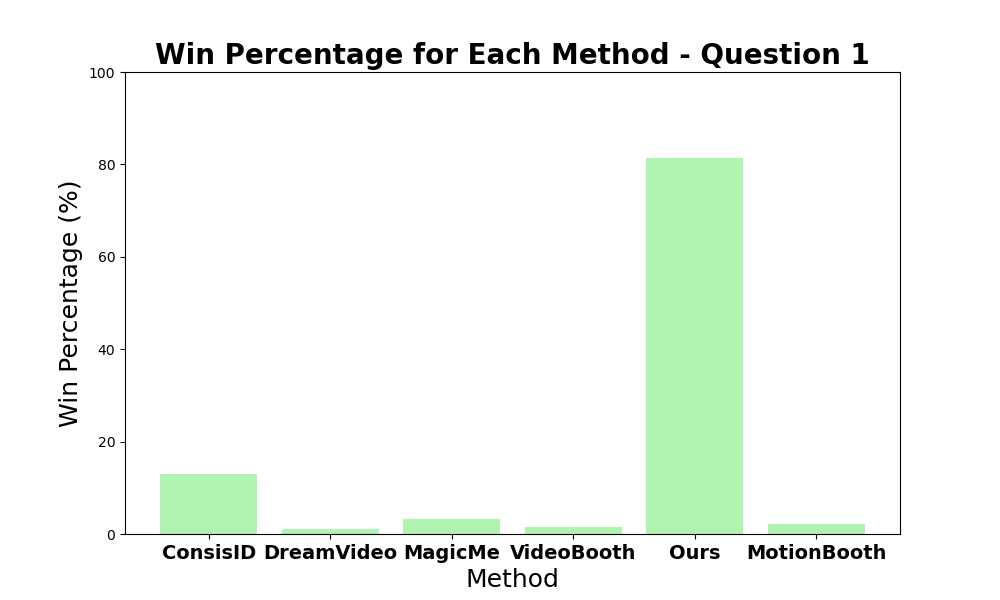}
    \caption{Statistics for the first question on identity preservation. }
  \end{subfigure}\hspace{8pt}
  \begin{subfigure}{0.32\linewidth}
    \centering
    \includegraphics[width=\linewidth]{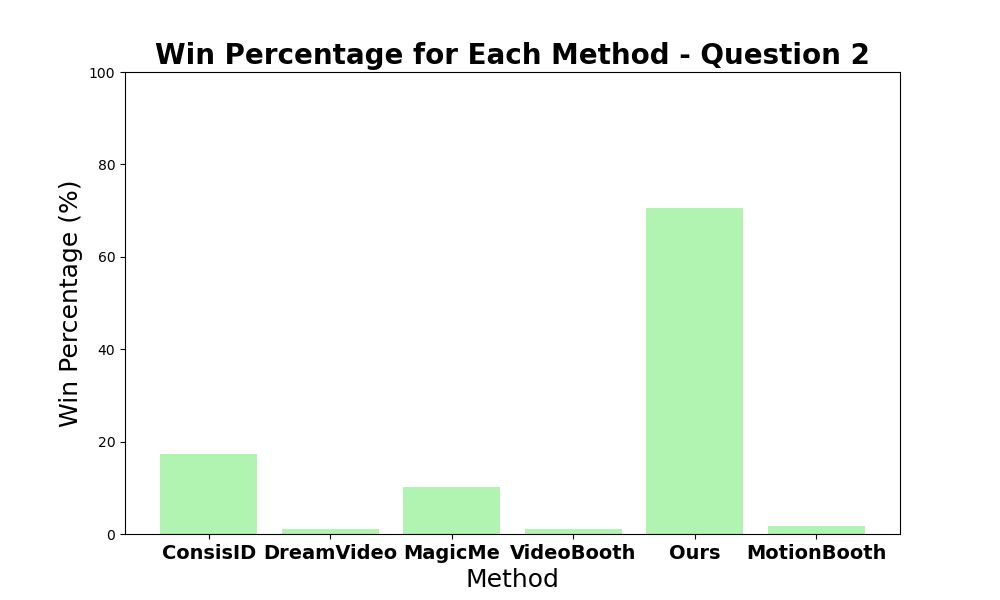}
    \caption{Statistics for the second question on natural integration of human face. }
  \end{subfigure}\hspace{8pt}
  \begin{subfigure}{0.32\linewidth}
    \centering
    \includegraphics[width=\linewidth]{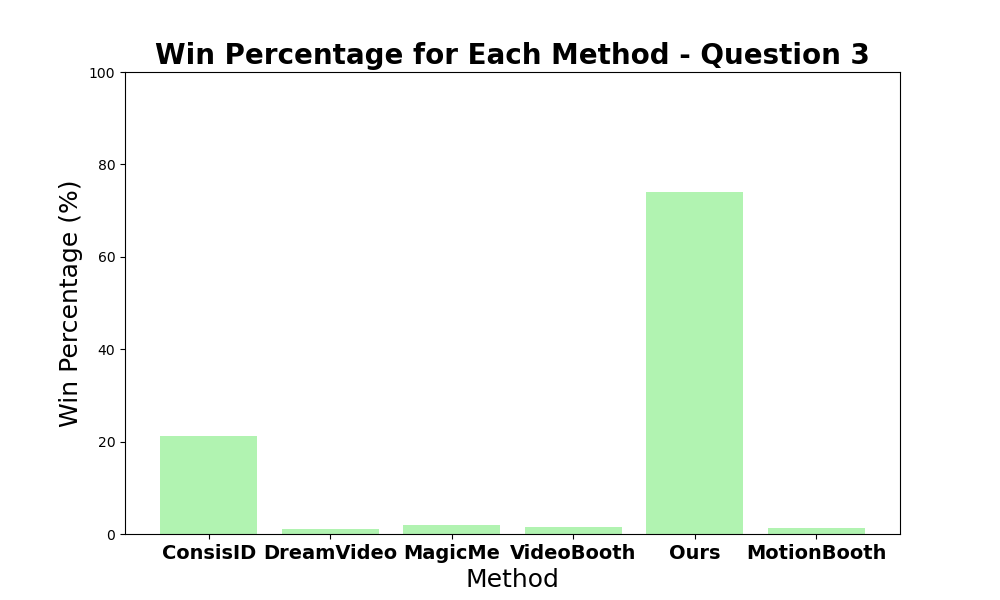}
    \caption{Statistics for the third question on text video alignment. }
  \end{subfigure}

  \caption{Screenshots and statistics of the user-study questionnaire on comparison with baselines in text-to-video customization.
  Our method secures the highest user-preference share in all questions. }
  \label{fig:supp_user_study_comparison_with_baselines}
\end{figure*}

Our method was selected as the best in 81.3\% of responses for multi-view identity preservation, 70.5\% for facial realism, and 74.1\% for text-video alignment. A screenshot and detailed statistics of the user study are shown in~\cref{fig:supp_user_study_comparison_with_baselines}.

\subsection{User study on the effect of joint-subject data} \label{app: user_study_joint_subject_data}

We conducted a user study across 60 prompts with 18 participants to evaluate the impact of joint-subject data on multi-subject video generation. Participants were asked to select the video that best preserved subject identity and depicted the most natural interaction between the subjects. The evaluation questions were:

\begin{itemize}
    \item \textbf{Multi-view identity preservation:} ``Which video best preserves the identity of the subject in the reference images?''
    \item \textbf{Natural multi-subject interaction:} ``In the videos, two people interact with each other. Which video depicts the most natural and realistic interaction between them?''
\end{itemize}

\begin{figure*}[htp]
  \centering
  \begin{subfigure}{\linewidth}
    \centering
    \includegraphics[width=\linewidth]{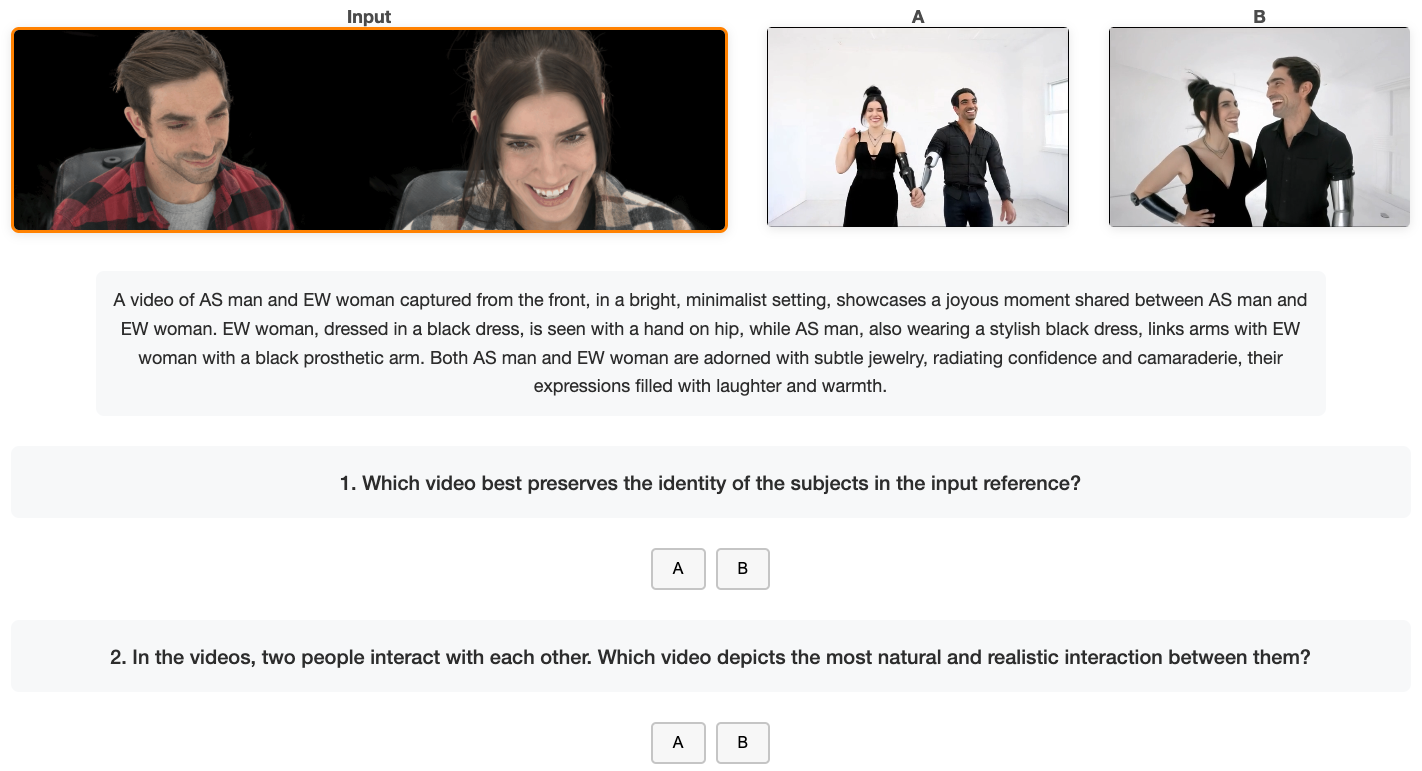}
    \caption{User-study interface and questions for the effects of joint-subject data.}
  \end{subfigure}
  \vspace{6pt} 

  \begin{subfigure}{0.48\linewidth}
    \centering
    \includegraphics[width=\linewidth]{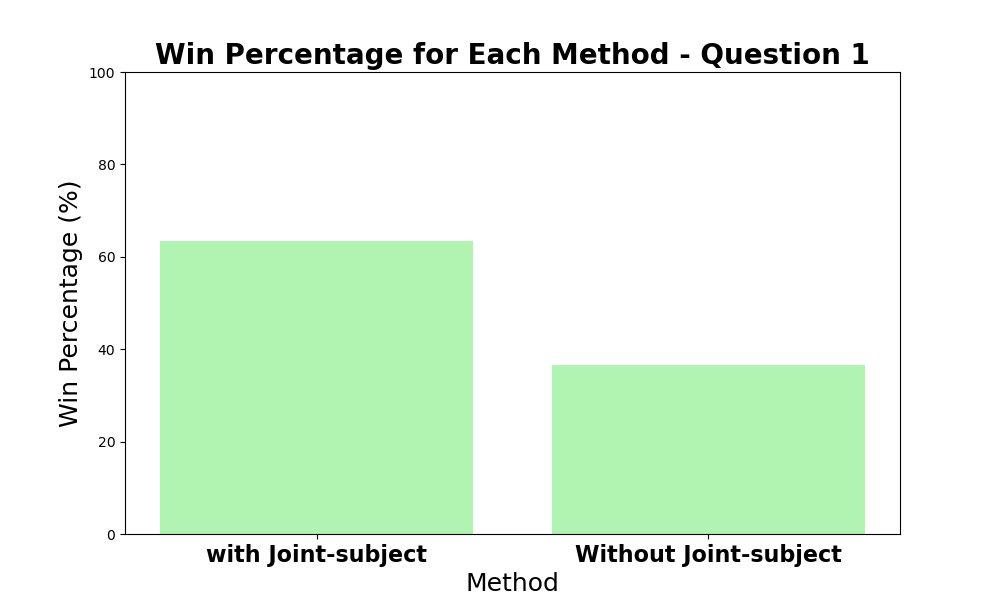}
    \caption{Statistics for the first question on identity preservation. }
  \end{subfigure}\hspace{8pt}
  \begin{subfigure}{0.48\linewidth}
    \centering
    \includegraphics[width=\linewidth]{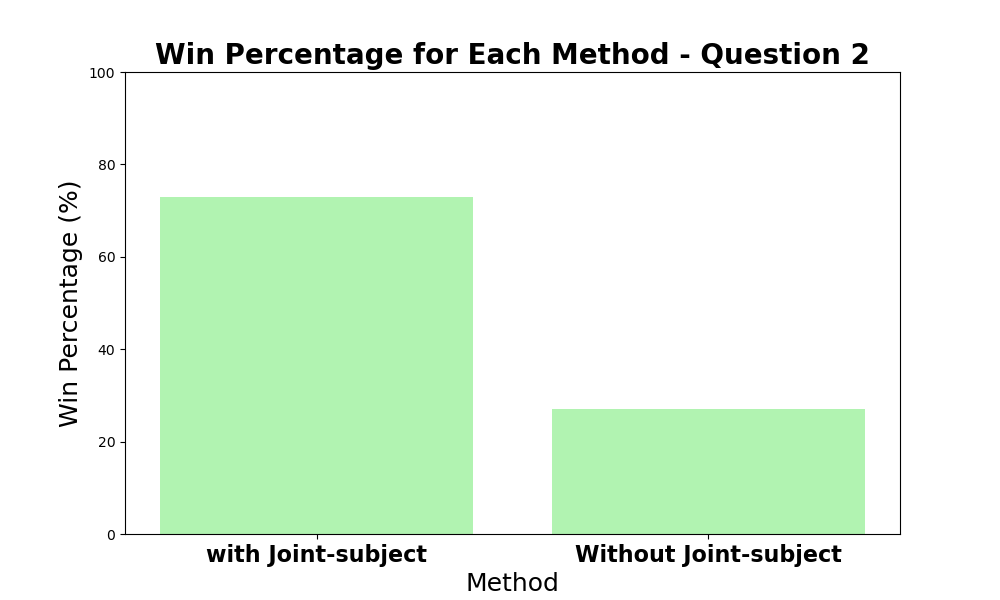}
    \caption{Statistics for the second question on natural human interaction. }
  \end{subfigure}\hspace{8pt}
  \caption{Screenshots and statistics of the user-study questionnaire on the effects of the joint-subject data in multi-subject generation 
  Using jointly-subject data secures the higer user-preference share in both questions. }
  \label{fig:supp_user_study_joint_subject_data}
\end{figure*}

The results show that the model trained with joint-subject data was preferred in 63.4\% of cases for identity preservation and 72.9\% of cases for natural interaction, highlighting the importance of joint-subject examples for improving multi-subject interaction realism. A screenshot and detailed statistics of the user study are shown in~\cref{fig:supp_user_study_joint_subject_data}.

\subsection{User study on the effect of relit data}

To evaluate the impact of incorporating relit 4DGS videos into the training data—aimed at enhancing lighting realism and variability—we conducted an ablation study comparing models trained with and without relit data. In a user study, 18 participants viewed pairs of videos generated by these two models across 60 prompts and selected the video with more realistic lighting and better identity preservation. The evaluation questions were:

\begin{itemize}
    \item \textbf{Multi-view identity preservation:} ``Which video best preserves the identity of the subject in the reference images?''
    \item \textbf{Natural lighting:} ``In some videos, the lighting appears uniformly flat, while in others, there is more variation in brightness and shadows, making the scene look more natural. Which video has the most realistic lighting with noticeable variations instead of flat illumination?''
\end{itemize}

\begin{figure*}[htp!]
  \centering
  \begin{subfigure}{0.7\linewidth}
    \centering
    \includegraphics[width=\linewidth]{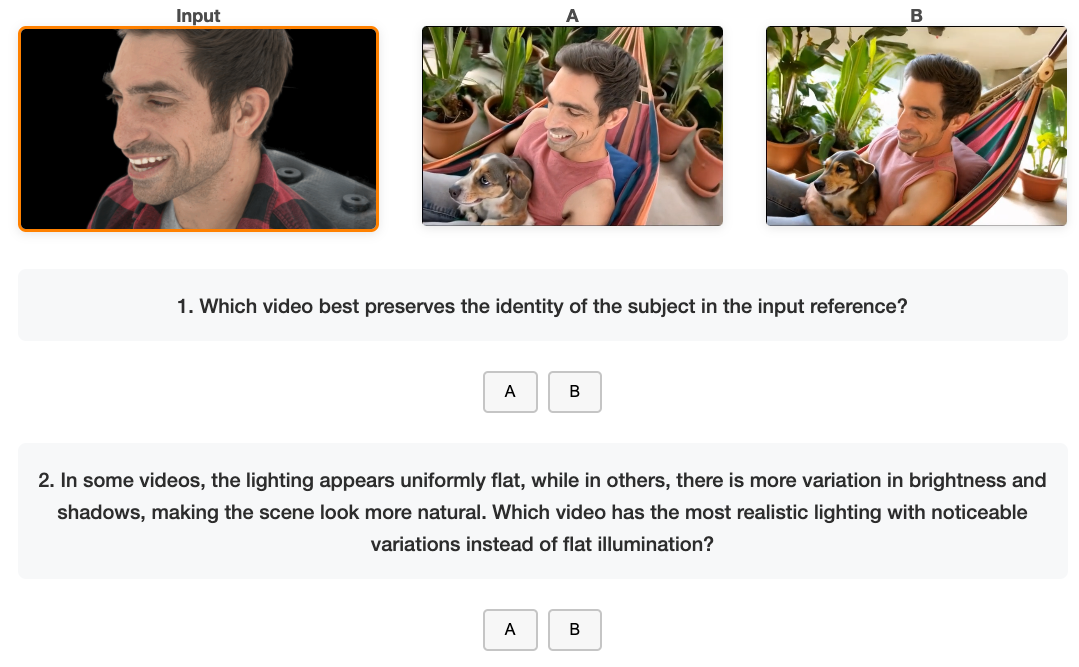}
    \caption{User-study interface and questions for the effects of relit data.}
  \end{subfigure}
  \vspace{6pt} 

  \begin{subfigure}{0.48\linewidth}
    \centering
    \includegraphics[width=\linewidth]{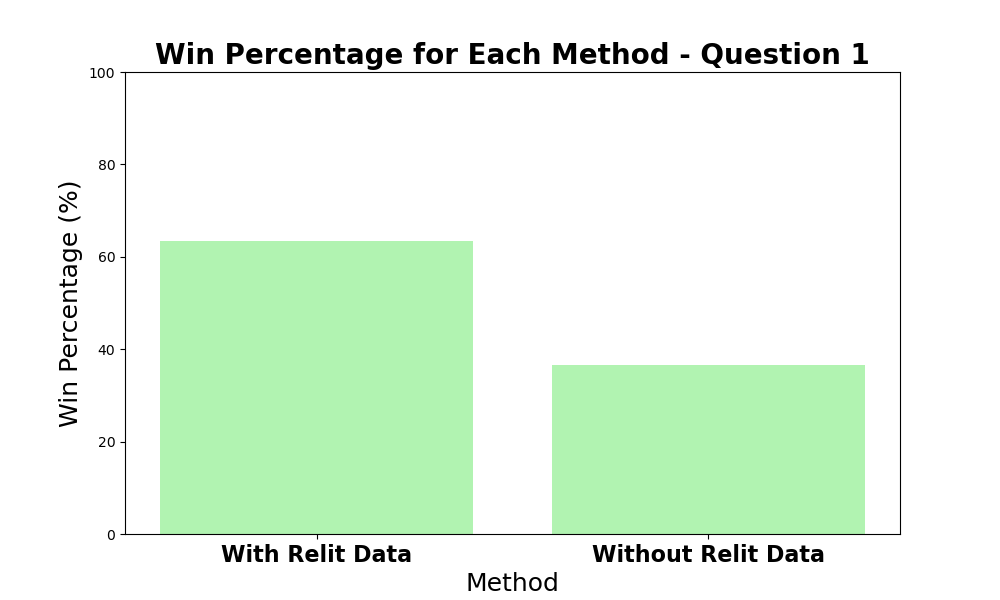}
    \caption{Statistics for the first question on identity preservation. }
  \end{subfigure}\hspace{8pt}
  \begin{subfigure}{0.48\linewidth}
    \centering
    \includegraphics[width=\linewidth]{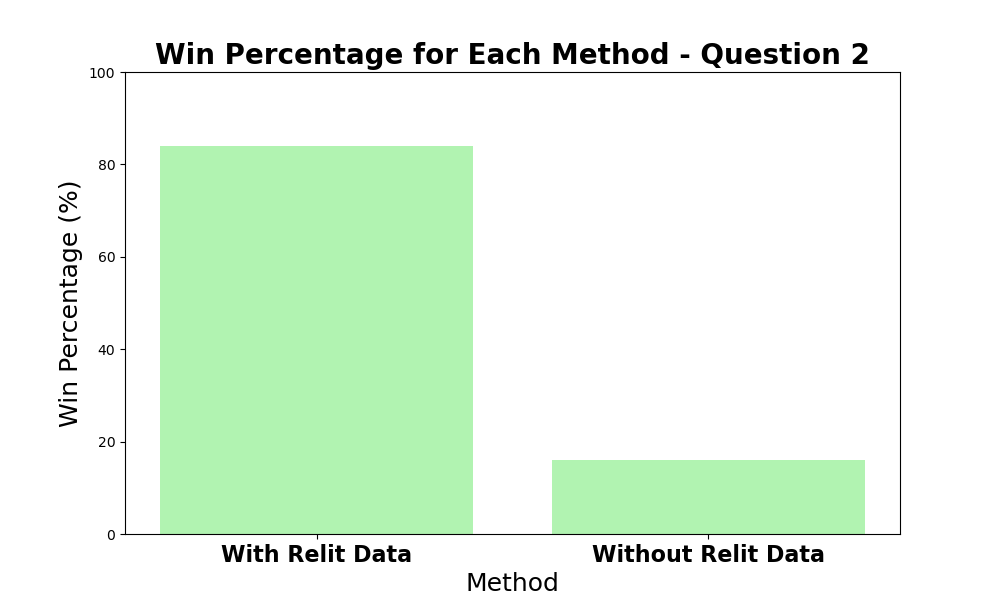}
    \caption{Statistics for the second question on realistic lighting. }
  \end{subfigure}\hspace{8pt}
  \caption{Screenshots and statistics of the user-study questionnaire on the effects of the relit data data in generating videos with realistic and diverse lighting.
  Adding relit data secures the higer user-preference share in both questions. }
  \label{fig:supp_user_study_relit_data}
\end{figure*}

The results show that the model trained with relit data was preferred in 83.9\% of cases for lighting realism and 63.0\% of cases for identity preservation, indicating that incorporating relit data not only improves lighting quality but also benefits identity consistency. A screenshot and detailed statistics of the user study are shown in~\cref{fig:supp_user_study_relit_data}.

\subsection{User study on customizing image-to-video model}

We conducted a user study with 18 participants across 60 prompts to evaluate the effect of customizing the image-to-video (I2V) model on identity preservation. For each prompt, participants were shown a pair of videos—one generated by a non-customized I2V model and the other by a customized I2V model—and asked the following question:

\begin{itemize}
    \item \textbf{Multi-view identity preservation:} ``Which video best preserves the identity of the subject in the reference images?''
\end{itemize}

\begin{figure*}[htp]
  \centering
  \begin{subfigure}{\linewidth}
    \centering
    \includegraphics[width=\linewidth]{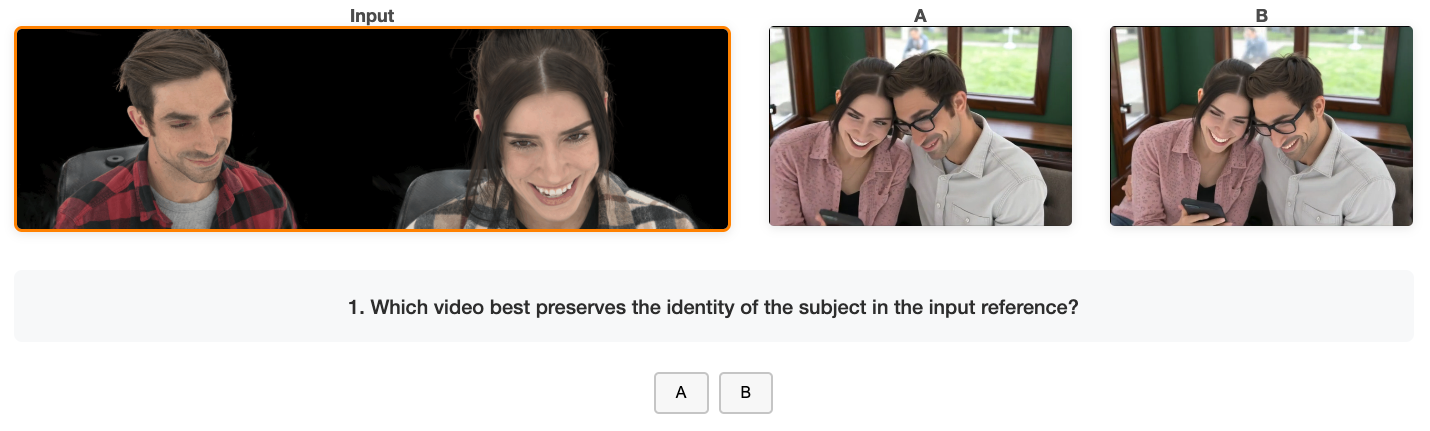}
    \caption{User-study interface and questions for the effects of customizing image-to-video model.}
  \end{subfigure}
  \vspace{6pt} 

  \begin{subfigure}{0.48\linewidth}
    \centering
    \includegraphics[width=\linewidth]{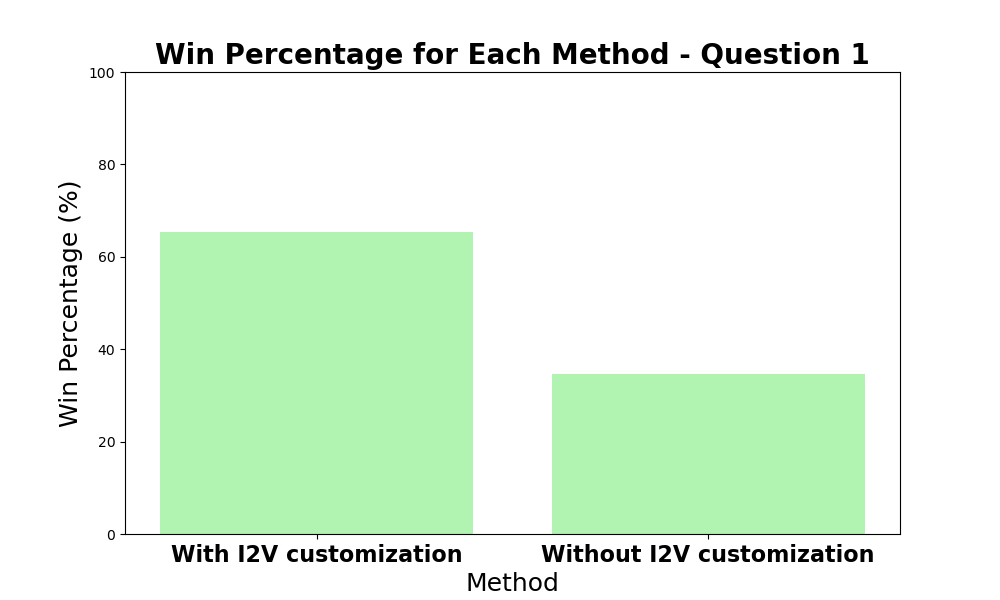}
    \caption{Statistics for the question on identity preservation. }
  \end{subfigure}\hspace{8pt}
  \caption{Screenshots and statistics of the user-study questionnaire on the effects of customizing image-to-video model in preserving the identity of the subjects.
  Customizing image-to-video model secures the higher user-preference share in identity preservation. }
  \label{fig:supp_user_study_I2V_customization}
\end{figure*}

The results show that 65.4\% of participants preferred the videos generated by the customized I2V model, highlighting the importance of customizing the I2V model for achieving high-quality, identity-consistent video generation. A screenshot and detailed statistics of the user study are shown in~\cref{fig:supp_user_study_I2V_customization}.

\end{document}